\documentclass[sn-mathphys]{sn-jnl}

\usepackage{amsmath}
\usepackage{booktabs}
\usepackage{adjustbox}
\usepackage{array}
\usepackage{amssymb,amsfonts}
\usepackage{graphicx}
\usepackage{caption,subcaption}
\usepackage{multirow}
\usepackage{textcomp}
\usepackage{wrapfig}
\usepackage{color}
\newcommand{\JY}[1]{\textcolor{black}{#1}}

\renewcommand{\footnote}{*}

\usepackage[capitalize]{cleveref}
\newcommand\n{\\\hline}
\usepackage{gensymb}
\usepackage{xurl}

\jyear{2023}%

\theoremstyle{thmstyleone}%
\theoremstyle{thmstyletwo}%

\theoremstyle{thmstylethree}%

\begin{document}

\title[ML for Leaf Disease Classification: Data, Techniques \& Apps]{Machine Learning for Leaf Disease Classification: Data, Techniques and Applications}


\author[1]{\fnm{Jianping} \sur{Yao}}\email{jianping.yao@utas.edu.au}

\author*[1]{\fnm{Son} \sur{N. Tran}}\email{sn.tran@utas.edu.au}

\author[2]{\fnm{Samantha} \sur{Sawyer}}\email{samantha.saywer@utas.edu.au}

\author[1]{\fnm{Saurabh} \sur{Garg}}\email{saurabh.garg@utas.edu.au}
\affil[1]{\orgdiv{School of Information and Communication Technology}, \orgname{University of Tasmania}, \orgaddress{\city{Launceston}, \postcode{7248}, \state{TAS}, \country{Australia}}} 

\affil[2]{\orgdiv{Tasmania Institute of Agriculture}, \orgname{University of Tasmania}, \orgaddress{\city{Prospect}, \postcode{7250}, \state{TAS}, \country{Australia}}} 


\abstract{The growing demand for sustainable development brings a series of information technologies to help agriculture production. Especially, the emergence of machine learning applications, a branch of artificial intelligence, has shown multiple breakthroughs which can enhance and revolutionize plant pathology approaches. In recent years, machine learning has been adopted for leaf disease classification in both academic research and industrial applications. Therefore, it is enormously beneficial for researchers, engineers, managers, and entrepreneurs to have a comprehensive view about the recent development of machine learning technologies and applications for leaf disease detection. This study will provide a survey in different aspects of the topic including data, techniques, and applications. The paper will start with publicly available datasets. After that, we summarize common machine learning techniques, including traditional (shallow) learning, deep learning,  and augmented learning. Finally, we discuss related applications. This paper would provide useful resources for future study and application of machine learning for smart agriculture in general and leaf disease classification in particular.}


\maketitle

\section{Introduction}

In recent years, Machine Learning (ML) has been emerging as a game changer in multiple aspects of life. In agriculture, machine learning has been widely used as an effective means of production, including but not limited to automatic harvesting machines, production estimation, pest control, weeds control, irrigation control, plant pathology (leaf disease classification), and fruit classification. Generally, diseases of a plant can react in different parts, such as its leaves, flowers and roots. Among them, plants’ leaf is one of the most dominant and pronounced parts. Because leaves can participate in providing the nutrients the plant needs to grow, which is the photosynthesis in leaves produces the chlorophyll from sunlight \cite{edssjs.15D0A2D220200101}. Some disease of leaves may cause their drop or wither, directly affecting the plant's yield and even survival. Furthermore, it will bring negative impacts, leading to crop productivity decrease, and production costs rise. In the past, farms generally rely on labour and experts for routine inspections and disease management. Their disadvantages are obvious. First, lots of manpower and costs are required. Second, labours need training and easily get fatigued on manual jobs. Third, it is difficult to detect leaf disease timely and on a large scale. Forth, diagnosis may be subjective due to human errors and bias. Thus, an effective leaf disease classification approach is the most basic need for plant cultivation. Fortunately, ML approaches have been recently emerging as a better solution compared to traditional methods, showing their effectiveness and ease of use in plant leaf pathology classification through plant leaf image analysing. Plant leaf images have several advantages. Datasets of leaves are relatively easy to collect, analyse and reproduce (e.g., using a camera). We can also extract useful features (e.g., species, healthy states, age, and disease categories), which would improve the quality and quantity of agricultural production. Therefore, efficient and timely identification and classification of plant diseases will be the key to remedying the loss of production.
Nowadays, with the introduction of precision agriculture (PA) or smart agriculture (SA) \cite{9418245, edssjs.4D1044AE20210101, edssjs.8410D5A320210101, 9238318, edssjs.15D0A2D220200101, 9742963}, ML technologies were researched and employed, especially in plant leaf pathology classification. Combine with Big Data and Internet of Things (IoT), ML can automatically detect plant leaf diseases as early as possible. Currently, the applications of ML have been deployed in various hardware and software, e.g., mobile phone applications\cite{9397001}, websites\cite{9342653} and smart glasses\cite{9182146}. With the increasing demand of ML in smart agriculture, a comprehensive survey on leaf disease classification will be beneficial to interested researchers and farmers. This paper would provide the research and industry communities with useful information on the available data and techniques, their advantages and weakness, and their applicability.

\JY{In recent years, there has been a growing interest in utilising machine learning for leaf disease classification. Several surveys have been conducted on this research topic; however, we have identified certain limitations within the reviewed works. The scope of the reviewed papers was often narrow, failing to encompass the broader concept of machine learning in leaf disease classification. Additionally, many of the reviewed papers were outdated, indicating a need for more up-to-date research in this area. Furthermore, a comprehensive review of available datasets for leaf disease classification is still lacking. It is also necessary to conduct a thorough review of the various machine learning approaches that have been employed. Currently, recent surveys have predominantly focused on emerging deep learning techniques, such as Convolutional Neural Networks (CNN). However, due to the diverse techniques and datasets used in each survey, it remains challenging to analyze and compare research outcomes. Moreover, while numerous software applications of machine learning for pathology, including leaf-disease analysis, have been developed recently, there is a lack of comprehensive review in this specific domain.}

This paper will provide a comprehensive view of current achievements and trends in the application of ML for leaf disease classification. Currently, leaf disease classification approaches can be categorised into traditional (shallow) ML, Deep Learning (DL) and Augmented Learning (AL). DL is a branch of ML and AL is a research topic, aiming to improve the effectiveness and usefulness of  ML approaches. In shallow learning, feature extraction plays an important role which, in many cases, requires experts' involvement, i.e. to engineer useful features. Deep learning, on the other hand, may reduce the cost of feature engineering as it can facilitate effective learning over a large amount of data. Although, data-hungry sometimes is an issue in deep learning, leaf images are sometimes easy to collect and farmers can help with disease annotation. However, to reduce the reliance on the labelled data, data augmentation methods have been taken to produce more training data and enhance the model robustness. Transfer learning is also a promising approach for this task, as it can reduce the need for leaf data by utilising pre-trained models from other tasks. As we can see, the keys to the success of ML approaches are the quality and quantity of data. Therefore, different from the other previous surveys, we discuss the availability and quality of public datasets and their suitability for evaluating ML models. 

 The organisation of the paper is as follows. In the next section, we will explain how we collect and analyse related literature. Section \ref{sec:rwork} will discuss the gaps in existing review and survey papers. After that, Section \ref {sec:Data_Sets} presents the available public datasets for leaf disease classification. This would help researchers to find, apply, and evaluate their ideas quickly. In Section \ref{sec:Approaches}, we categorise and compare machine learning approaches, by dividing them into three main groups: traditional (shallow) ML approaches, deep learning (DL),  and transfer learning (TL). In Section \ref{sec:apps}, we present related applications available for leaf disease classification in real-life. Finally, Section \ref{sec:concl} will summarise our findings and discuss the potential directions for future work on this research topic. This paper aims to provide some useful resources for the study and application of leaf disease classification with machine learning. 

\section{Methodology}
\label{sec:methodology}
\begin{figure*}[h]
    \centering
    \includegraphics[width=0.7\textwidth]{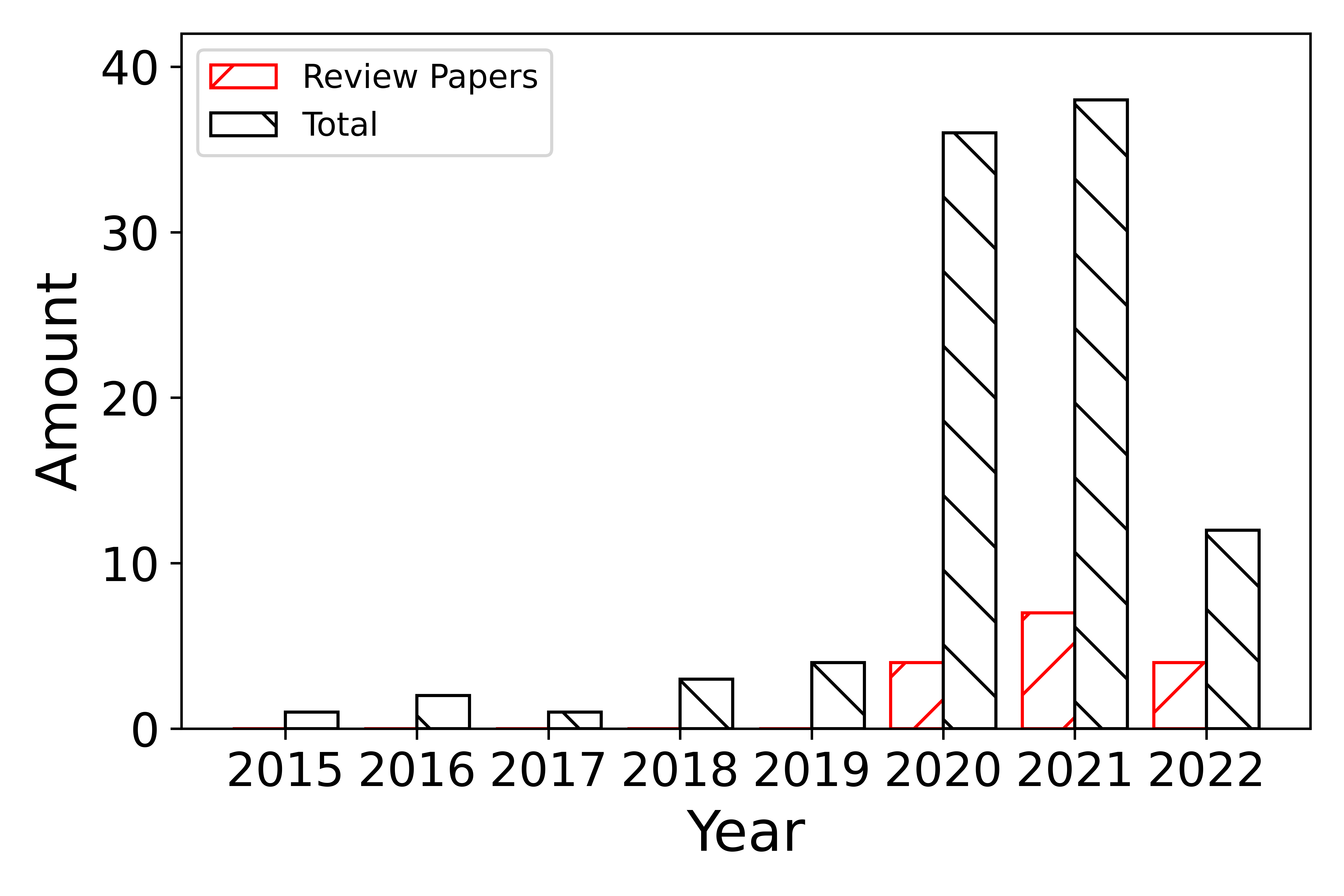}
    \centering
    \caption{\JY{The Amount \& Years of Referenced Articles. The red color indicates the number of review paper on leaf disease classification.}}
    \label{fig:amount_year_2}
\end{figure*}

This study was researched through a series of well-known databases, including EBSCO host, Scopus and Google Scholar. The search keywords were including “leaf disease”, “plant disease”, “machine learning”, “deep learning”, “classification”, “detection”, etc. In this review, we firstly focus on quality papers by filtering them using 3 metrics: (1) number of citations; (2) rank of the published venues (Q1 for journals, and rank CORE A/A* for conferences); and (3) relevance. In addition, beyond the criteria, we also studied as many relevant articles as we could find to avoid the issue of omission. As shown in Table \ref{Methodology} and Fig.\ref{fig:amount_year_2}, the academic articles referenced mainly focus on the recent years (2015 -2022). 
\JY{In Table \ref{Methodology}, the review papers are denoted with asterisks. Out of the total papers published from 2020 to 2022, there are 15 review papers and 71 technical papers.}
In Fig.\ref{fig:amount_year_2}, the amount of papers shows an increasing trend year by year, which reflects the growing interest in plant leaf detection and classification. \JY{As we can see, the number of papers increases substantially in recent years, showing a growing interest in this topic.}

	\begin{table}[h!]
		\centering
			\begin{minipage}{\textwidth}
				\caption{The publication years of Referenced Academic Articles. "Tech" column shows the number of technical papers, while "Review" column shows the number of review papers. The total number of papers we study in each year is in "Total" column.}\label{Methodology}%
				\begin{tabular}{|c|m{6cm}|c|c|c|}
					\hline
					\multirow{1}{*}{Year} & \multicolumn{1}{c|}{Paper} & Tech  & Review  & Total \\
					\hline
					\hline
					2022 & \cite{9725870}\footnote, \cite{9742963}\footnote, \cite{524942744}, \cite{9752495}, \cite{520693437}, \cite{9753846}, \cite{9753629}, \cite{9823799}, \cite{531183386}\footnote, \cite{517091827}, \cite{523682058}\footnote, \cite{2b1bee20d14245e49c23b3dfe45cded2} &  8 & 4 & 12\\
					\hline					
					2021 & \cite{9388488}\footnote, \cite{SUJATHA2021103615}, \cite{9418245}, \cite{9422499},\cite{9418680}, \cite{JEPKOECH2021107142}, \cite{9331214}, \cite{9396023}\footnote, \cite{9397001}, \cite{9408806}, \cite{15100606920210401}, \cite{9393311}, \cite{15107730820210601}, \cite{9418013}, \cite{sharma2021detection}, \cite{15100606020210401}, \cite{edssjs.30A9493B20210101}, \cite{9399342}\footnote, \cite{edssjs.8410D5A320210101}, \cite{edssjs.4D1044AE20210101}, \cite{edsdoj.832f5b1033dd4a7599ac9b755b17c7b920210601}, \cite{15142389920210101}, \cite{S004579062100047120210301}, \cite{a_15100615120210401}, \cite{electronics10121388}, \cite{9568324}\footnote, \cite{14844519620210101}, \cite{Kathiresan_2021}, \cite{edsdoj.8ed1e30391ba44ac92cb5f6e09b0fdf420210401}, 
					\cite{exsy.12885}, \cite{15142376220210101}\footnote, \cite{9544640}\footnote, \cite{419621445}, \cite{9718941}, \cite{9675567}, \cite{9509632}, \cite{9587775}\footnote, \cite{agriengineering3030032}         &31 & 7 & 38 \\
					\hline
					2020 &\cite{10.3389/fpls.2020.00751}, \cite{LU2020105760}\footnote, \cite{Singh_2020}, \cite{9355991}, \cite{9362812}, \cite{9076371},  \cite{9130019}, \cite{9291694}, \cite{9261801}, \cite{9293866}, \cite{9077134}, \cite{9350413}, \cite{9137986}, \cite{9155585}, \cite{9250911}, \cite{9250885}, \cite{9278067}, \cite{9277379}, \cite{9392051}, \cite{9342653}, \cite{9137805}, \cite{9198326}, \cite{9142988}, \cite{9138030}, \cite{9342729}, \cite{9212816}, \cite{9231174}, \cite{9182128}, \cite{9182146}, \cite{9238318}\footnote, \cite{9210294}, \cite{9057889}, \cite{S187705092030690620200101}, \cite{t_14770449620201201}, \cite{9243434}\footnote, \cite{edssjs.15D0A2D220200101}\footnote &32 & 4 & 36 \\
					\hline
					2019 - 2015 &\cite{S004579061930002320190601}, \cite{8974752}, \cite{8944556}, \cite{mwebaze2019icassava}, \cite{8374024}, \cite{8566635},\cite{7746160}, \cite{DBLP:journals/corr/HughesS15}, \cite{hughes2016open}, \cite{2997017820180703}, \cite{IDT-170301} &11 & 0 & 11 \\
					\hline
				\end{tabular} %
				
				\footnote{Review papers}
				
		\end{minipage}
		
\end{table}

\section{Related Work}
\label{sec:rwork}

\begin{table}[h!]
	\begin{center}
		\begin{minipage}{\textwidth}
			\caption{Recent Review Papers}\label{Recent_Review_Papers}%
					\begin{tabular}{@{}ccp{0.4\textwidth}p{0.4\textwidth}@{}}					
						\toprule
						Paper & Year & Strength & Limitation \\
						\midrule
						\cite{9725870} & 2022 & Detailed plant leaf disease introduction & Small number of articles, image processing with ML (3 articles), DL (5 articles) and SI (5 articles)	\\	
						\cite{9742963} & 2022 & Focus on potato leaf diseases  & Just 8 articles about the potato leaf diseases classification results.   \\
						\cite{531183386} & 2022 & List the factors may produce plant disease  & Did not list available public datasets \\
						\cite{523682058}  &2022  & Detailed pre-processing, feature extraction \& classifier analyses & Did not list available public datasets\\
						\cite{9396023}  & 2021 &  Concluded that multi-layer CNN performance is better than shallow ML & The number of papers they reviewed is relatively small (17 papers) \\
						\cite{9568324}    & 2021  & Focused datasets, categories, DL methods and average accuracy  & Most of papers worked on Plant Village, The number of papers was too small (10 papers) \\
						\cite{9388488} & 2021 &  Detailed analysed 45 recent papers, pointed out the deficiency of SL compared to DL and feature extraction may be unnecessary to DL & Different accuracy from different datasets, there may be a lack of benchmark datasets \\
						\cite{9399342} & 2021 &  Analysed the weaknesses of SL, affirmed the superiority of DL and introduced transfer learning. Besides, reviewed the common visualization techniques for explainability.  & Because of the lack of available data most of the models had poor robustness;  only suitable for special species and leaves \\	
						\cite{15142376220210101} & 2021& Reviewed recent developments of DL (CNN, DNN and TL) & A few  articles and studies in  general\\		
						\cite{9544640} & 2021 & Some summaries of plant pathology      &        A few articles about Leaf disease classification and a small number of publicly available datasets \\
						\cite{9587775} & 2021 & Detailed analyses of image segmentation, feature extraction and classification stages     &     Most of the work is about shallow ML \\
						\cite{9243434}   & 2020  & Focused on disease recognition, analysed crop types, disease Types, test set percentages, common techniques and classification accuracy, and common techniques and their accuracy  & Most of the papers are before 2017 and the focus is on detection tasks (disease/not disease)\\
						\cite{edssjs.15D0A2D220200101} & 2020 & Detailed analysed related PA concepts and technologies (esp. Colour Space Models \& Feature Extraction) and introduced the common diseases  & Most of the listed datasets are private and access unavailable\\
						\cite{LU2020105760} & 2020 & Analysed many publicly available datasets (15 weed control, 10 fruit classification and 9 others) & Only the Maize Leaf (NLB) Dataset is for leaf disease classification \\
						\cite{9238318} &2020 & Reviewed 26 papers’ datasets, methods \& results and confirmed DL’s efficiency &  Did not list available public datasets\\		
					 					
						\bottomrule
					\end{tabular} %
			\end{minipage}
		\end{center}
	\end{table}

	As the interest in leaf disease classification with machine learning has been increasing recently, there are several surveys related to this research topic. In this section, we analyse recent review papers about leaf disease classification or classification. Table \ref{Recent_Review_Papers} shows their study and the gaps they left behind. As we can see, the previous surveys focused on different aspects of leaf disease classification, shedding light on some key areas in the research topic but a comprehensive study is still missing. 
	
	First, we found that many related works have a shadow scope for their study. The number of papers for review is not adequate to cover the broad concept of ML in leaf disease and many papers used in the reviews are not up-to-date. For example, In \cite{9396023, 9568324}, no more than 20 articles are selected from Google Scholar for their study. Another survey paper \cite{9243434}, published in 2020, analyse articles all before 2017. \cite{9238318} analysed 26 academic papers about leaf disease detection and classification from 2015 to 2020. \cite{9388488} surveyed more than 45 academic papers about plant disease detection and classification from 2017 to 2020. \cite{9544640} has 12 papers focusing on deep learning techniques only. In \cite{531183386}, they review shallow ML (10 articles) and DL (20 articles, including TL). \cite{9725870} surveyed about image processing with ML (3 articles), DL (5 articles) and SI (5 articles). \cite{9742963} just includes 8 articles about the potato leaf disease classification results. In a recent survey \cite{523682058}, 179 papers have been studied, however, there are only 12 articles are from recent years (2020-2022) and not all of them are about leaf disease classification (the survey also covers plant species classification). Different from it, our paper focuses on more recent studies.

	Second, we found that a comprehensive review about the available datasets of leaf disease classification is still missing. Many researchers already noticed that the primary obstacle in this research topic is the availability of datasets \cite{edssjs.15D0A2D220200101, 531183386, 9399342,9544640}. For example, \cite{LU2020105760} surveyed 34 agricultural datasets, however, there is only one dataset, the Maize Leaf (NLB) \cite{2997017820180703}, which is related to leaf diseases. Unfortunately, many datasets introduced in related work listed here are private \cite{edssjs.15D0A2D220200101, 9238318, 531183386, 9544640}. Plant Village is one of the most popular public datasets \cite{9396023, 9568324, 9544640, 531183386, 9725870, 9742963, 9544640, 523682058}. This dataset is useful for the scientific research purpose, however, there are some pitfalls due to its laboratory-condition. In, \cite{531183386, 9544640, 9725870}, the authors expressed the importance of real-field datasets. In another research, a combination of public (55\% based on Plant Village) and private data (25\% ) is used \cite{523682058}. Recently, more calls on the availability of leaf disease data to bring greater benefits to both scientific and industrial communities  \cite{9544640}.
	
	Third, there are many different machine-learning approaches, and they need to be reviewed thoroughly. Early survey studies focus on traditional (shallow) approaches such as Artificial Neural Networks (ANN), Support Vector Machine (SVM), AdaBoost, KNN, Decision Tree, Naïve Bayes (NB) \cite{9396023, 9243434, 9388488, 9742963, 9587775, 523682058}. In these approaches, data pre-processing and feature engineering are usually needed \cite{9396023, 9399342, 9243434, 9587775}. Feature engineering is an important step to extract the features of images as inputs for ML models \cite{9388488}. Normally, hand-crafted features will be extracted which requires the involvement of humans, i.e. domain experts to define useful features. For feature extraction, there exists a wide range of methods, including Local Binary Patterns (LBPs) Histogram, Speeded Up Robust Features (SURF), Scale Invariant and Feature Transformation (SIFT),  Gabor Energy Filtering, Principal Component Analysis (PCA), Linear Discriminant Analysis (LDA), Generalized Extreme Value (GEV) Distribution and Johnson SB Distribution \cite{523682058}. 
	
Recent surveys have revolved around new techniques, including deep learning, such as, CNN\cite{15142376220210101, 9388488, 9742963, 9544640, 523682058}, AlexNet, GoogLeNet, and VGGNet \cite{9388488, 9742963, 523682058}, Pooling Dilated CNNs \cite{9396023}. Recently, traditional (shallow) approach has been replaced by deep learning methods \cite{S004579061930002320190601}, as it may cause side effects (\cite{sharma2021detection, 9231174}) due to human errors/biases during feature engineering step. A number of experimental results showed that DL is a powerful and useful way to detect and classify leaf diseases \cite{9238318, 9399342, 9396023, 9742963, 15142376220210101, 9388488}. DL technologies are relatively user-friendly, can extract image features and classify plant diseases automatically \cite{9399342}. 
\JY{For example, the higher accuracy of DL compared to the traditional (shallow) approach was demonstrated by \cite{531183386}. They found that DL models, with and without pre-training, achieved average accuracies of 99.64\% and 98.64\% respectively, surpassing the 95.71\% accuracy of the traditional approach.}
For improvement, recent studies enhance the performance of machine learning models, especially deep learning, with supplementary techniques, such as segmentation \cite{531183386, 9725870, 523682058}, data augmentation \cite{9399342}, and transfer learning \cite{9399342, 15142376220210101, 9544640}, or combination of traditional and deep learning \cite{9388488}. \cite{9399342} claimed that transfer learning would be the most effective method to boost the robustness of CNN classifiers. \cite{9388488} employed a combination of different segmentation algorithms to extract better features of the images.

As we can see, each survey focuses on a different set of techniques and data based on various timelines. This makes it difficult to analyse and compare the research outcomes. Moreover, many software applications of ML for pathology, including leaf-disease analysis, have been developed recently and there is a lack of a review in this aspect. In this paper, we will address the limitations above by providing a comprehensive review of recent studies, public datasets, machine learning techniques, and real-life applications of machine learning in leaf disease classification.

\section{Datasets}
\label{sec:Data_Sets}
\begin{figure*}[h]
\centering
\includegraphics[width=\textwidth]{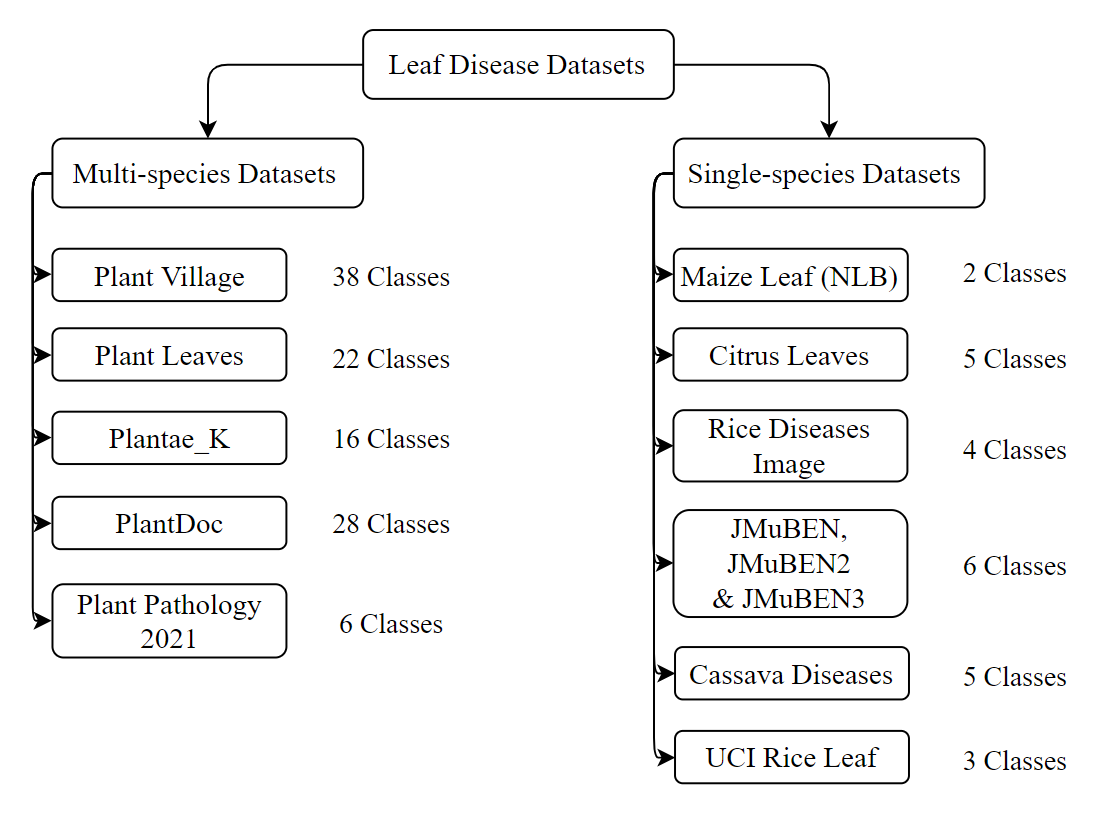}
\centering
\caption{The Structure of Public Datasets  } 
\label{fig:Structure_of_Datasets}
\end{figure*}

Data plays a critical role in modern AI, especially in the emergence of deep learning techniques recently. The quantity and quality of training data will improve the performance of large models used in deep learning \cite{ Goodfellow-et-al-2016}. In research and practice, the role of image datasets for computational vision tasks is self-evident. In \cite{edssjs.15D0A2D220200101}, a study showed that the foremost challenge for research is the lack of available datasets.  For leaf disease classification, in recent years, many researchers have devoted themselves to the collection of plant disease data for public use. Table \ref{Public_Dataset_Address} and Figure \ref{fig:Structure_of_Datasets}  show recent available public datasets about plant leaf diseases for computer vision research. In the table, the “Year” column represents the published year of a dataset. “Species” shows the number of plant species. The “Diseases” column lists the number of unique diseases. We also include a “Class” column to show the number of original classes in the dataset, as some datasets combine species and diseases as labels. We categorise the datasets into a multi-species group and a single-species group according to their species diversity.

\begin{table}[h]
\begin{center}
\begin{minipage}{\textwidth}
\caption{Public Leaf Disease Datasets}\label{Public_Dataset_Address}%
\resizebox{\textwidth}{!}{%
\begin{tabular}{@{}lccccl@{}}
\toprule
Dataset & Year & Species & Disease & Class & Link \\
\midrule
Plant Village   & 2016 & 14 & 22 & 38 & \url{https://data.mendeley.com/datasets/tywbtsjrjv/1}\\
plant leaves & 2019 & 12 & 22 &22 & \url{https://data.mendeley.com/datasets/hb74ynkjcn/1} \\
Plantae\_k &  2019 & 8 & 9 & 16 & \url{https://data.mendeley.com/datasets/t6j2h22jpx/1}  \\
PlantDoc & 2020 & 13 & 17 &28 & \url{https://github.com/pratikkayal/PlantDoc-Dataset} \\
Plant Pathology 2021 - FGVC8  & 2021 & 1 & 6 & 6 & \url{https://www.kaggle.com/c/plant-pathology-2021-fgvc8/overview}  \\
Maize Leaf (NLB) & 2018 & 1 & 2 & 2 & \url{https://osf.io/p67rz/}\\
Citrus Leaves & 2019 & 1 & 5 & 5 & \url{https://data.mendeley.com/datasets/3f83gxmv57/2}\\
Rice Diseases Image Dataset & 2019 & 1 & 4 & 4 & \url{https://www.kaggle.com/minhhuy2810/rice-diseases-image-dataset} \\
\begin{tabular}{@{}l@{}} JMuBEN \\ JMuBEN2  \end{tabular} Arabica Coffee Leaf Images  & 2021 & 1 & \begin{tabular}{@{}l@{}} 3 \\ 2  \end{tabular} & 5 & \begin{tabular}{@{}l@{}} \url{https://data.mendeley.com/datasets/t2r6rszp5c/1} \\ \url{https://data.mendeley.com/datasets/tgv3zb82nd/1} \end{tabular}  \\
 JMuBEN3 Sweet Potato Leaf Spot & 2021 &  1 &  1  &  1  & \url{https://data.mendeley.com/datasets/jjn4ht687d/3}  \\
 Cassava Diseases & 2019 & 1 & 5 & 5 & \url{https://www.kaggle.com/c/cassava-disease/data}  \\
 UCI Rice Leaf Diseases & 2017 & 1 & 3 & 3 &  \url{https://archive.ics.uci.edu/ml/datasets/Rice+Leaf+Diseases} \\
\bottomrule
\end{tabular} %
}
\end{minipage}
\end{center}
\end{table}

\subsection{Single-species Datasets}
A single-species dataset is specific to one plant species. It can be used in the detection, classification or severity assessment of a specialised plant.

\noindent\textbf{Plant Pathology 2021 - FGVC8 Dataset.} Plant Pathology 2021-FGVC8 is an apple leaf disease image dataset of a Kaggle challenge competition. It is a part of the Fine-Grained Visual Categorization FGVC8 workshop at the Computer Vision and Pattern Recognition Conference (CVPR) 2021. This dataset is characterised by each leaf having 1 or several labels. It contains around 23,000 apple images, and six apple leaf health categories: "healthy", "complex", "rust", "frog eye leaf spot", "powdery mildew", and "scab". Among them, “complex” means a leaf is unhealthy but we are unable to identify an exact cause (disease). This dataset would be useful for multi-class apple leaf disease classification.

\noindent\textbf{Maize Leaf (NLB) Dataset}. The Maize Leaf (NLB) Dataset was collected through various shooting methods proposed by \cite{2997017820180703}. This includes hand cameras,  cameras on a 5 m boom, and cameras on a drone. The dataset has more than $18,222$ maize plant images with $105,735$ Northern leaf blight (NLB) lesions annotated by experts. This dataset is considered as the largest open dataset of single plant species at present, and will be helpful for maize disease classification and severity assessment.

\noindent\textbf{Citrus Dataset.} The Citrus dataset has two folders, 150 images of citrus fruits and 609 images of citrus Leaves, each folder has 5 categories (black spot, canker, greening, melanose, and healthy). All images were annotated by experts.

\noindent\textbf{Rice Diseases Image Dataset.} Rice Diseases Image Dataset has four categories of rice leaves: Brown Spot (523 images), Healthy (1488 images), Hispa (565 images) and Leaf Blast (779 images). The dataset has been studied in several works \cite{Kathiresan_2021, edsdoj.8ed1e30391ba44ac92cb5f6e09b0fdf420210401} for leaf disease classification.

\noindent\textbf{JMuBEN Datasets (JMuBEN, JMuBEN2, JMuBEN3).} This is a group of datasets (JMuBEN, JMuBEN2, JMuBEN3) that were released by the same authors \cite{JEPKOECH2021107142} and were all collected by a camera under plant pathologists’ guide. JMuBEN and JMuBEN2 are about Arabica coffee leaves that were taken from real coffee plantations. They can be combined into a larger dataset. JMuBEN has three categories: 7682 Cerscospora images, 8337 rust images and 6572 Phoma images. JMuBEN2 has two categories: 16,979 Miner images and 18,985 healthy images. JMuBEN3 is about sweet potato leaves which are all affected by leaf rust. It just has one category: 1383 Sweet potato leaf rust images. The JMuBEN3 dataset folder also contains a sweet potato leaf rust classification model code by the authors. Some images of JMuBEN and JMuBEN2 were augmented by rotation and flipping methods to increase dataset size and prevent the over-fitting issues \cite{JEPKOECH2021107142}. These datasets are useful for deep learning research and study.

\noindent\textbf{Cassava Disease Dataset.} Cassava disease dataset is from a Kaggle challenge competition as a part of the Fine-Grained Visual Categorization workshop (FGVC6) at CVPR 2019. It contains 1 healthy and 4 disease categories which are Cassava Brown Streak Disease (CBSD), Cassava Mosaic Disease (CMD), Cassava Bacterial Blight (CBB) and Cassava Green Mite (CGM). All images were collected by 200 farmers through small phones and annotated the labels by experts. The dataset has two parts, one is a training set (9,436 annotated images) and another is a test set (12,595 unlabeled images). In the dataset, the experts also scored the disease severity (from 1 to 5), however, the Kaggle did not include the scores \cite{mwebaze2019icassava}.

\noindent\textbf{UCI Rice Leaf Diseases Dataset}. UCI Rice Leaf diseases dataset aims to use for rice plant diseases detection and classification \cite{IDT-170301}. It has three disease categories: {Bacterial leaf blight}, Brown spot, and Leaf smut, and each category has 40 images. The limitation of it is the size is too small (120 images total). This can be useful for prototyping machine learning methods for quick testing but may not be suitable for deep learning approaches which require large amounts of data.

\subsection{Multi-species Datasets}
A multi-species dataset is composed of a variety of plant species, each has its own (overlapping) set of diseases. The datasets in this group can be used for the classification of species and classification of diseases.

\noindent\textbf{Plant Village Dataset.} Plant Village Dataset is currently the most widely used and popular public dataset for leaf disease classification. It has two versions,  an original version and a data augmentation version. The original dataset was published in 2016 by \cite{hughes2016open} with 54,305 leaf diseases or healthy images from 14 plant species (e.g., Apple, Blueberry, Cherry and Corn). Each species has 1-10 classes of related diseases or healthy (22 unique disease categories total). In the dataset folder, it has a total of 38 classes that combined species and diseases (e.g., Apple black rot), and one additional category of about 1143 background images (without leaf). The data augmentation version was released in 2019 by \cite{DBLP:journals/corr/HughesS15}, they used six data augmentation methods ( i.e. image flipping, Gamma correction, noise injection, PCA colour augmentation, rotation, and Scaling) to enhance the data. As a result, the original dataset had been increased from 54,305 to 61,486 images.

\noindent\textbf{Plant Leaves Dataset.} Plant Leaves dataset consists of 4502 images of healthy and unhealthy leaves divided into 22 categories by species and state of health. The images are in high-resolution JPG format. 12 tree types are  AlstoniaScholaris, Arjun, Bael, Basil, Chinar, Gauva, Jamun, Jatropha, Lemon, Mango, Pomegranate, and PongamiaPinnata. Notice that the Bael class only has diseased leaves and Basil only has healthy leaves.

\noindent\textbf{Plantae\_K Dataset. }
Plantae\_K dataset contains $2,153$ images of healthy and unhealthy plant leaves, divided into 16 categories by species and state of health (e.g., apple healthy and apple diseased). The images are in high-resolution JPG format. There are 8 fruit types in this dataset, including Apple, Apricot, Cherry, Cranberry, Grapes, Peach, Pear and Walnut.

\begin{figure*}[h]
	\centering
	\begin{subfigure}{0.44\textwidth}
		\centering
		\includegraphics[width=0.9\textwidth]{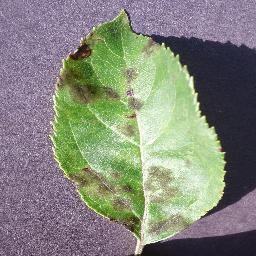}
		\caption{ Plant Village }
		\label{fig:plant_village_apple_scab}
	\end{subfigure}
	\begin{subfigure}{0.44\textwidth}
		\centering
		\includegraphics[width=0.9\textwidth]{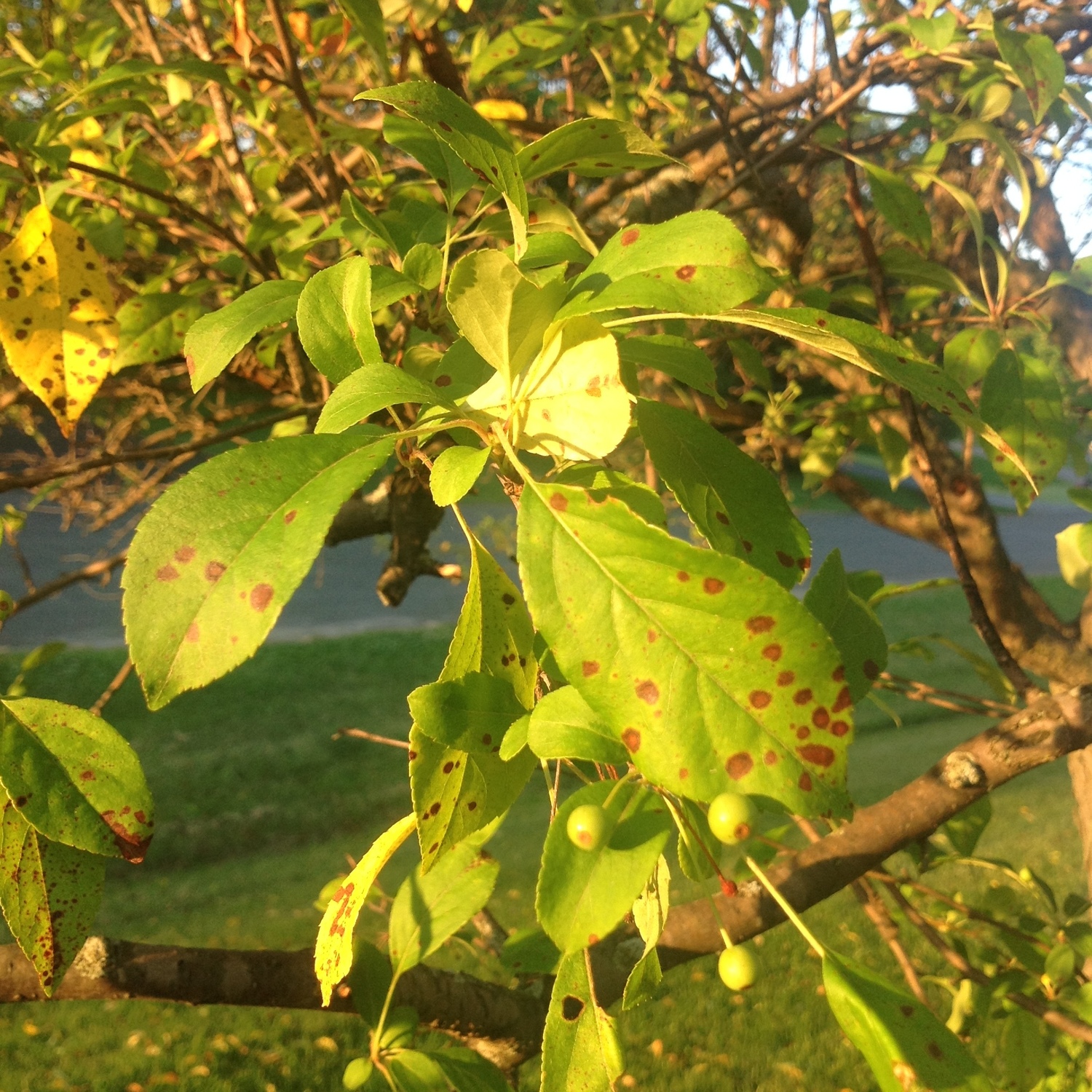}
		\centering
		\caption{PlantDoc}
		\label{fig:PlantDoc_apple_scab}
	\end{subfigure}
\caption{Apple Scab Leaf Samples}
\label{fig:Apple_Scab_Leaf}
\end{figure*}

\noindent\textbf{PlantDoc Dataset. }
Compared to Plant Village Dataset, the PlantDoc dataset aims to establish a real-field images dataset. \cite{Singh_2020} concerned that the images of Plant Village (e.g., \ref{fig:plant_village_apple_scab}) were all taken in laboratory setups and not in the real conditions of cultivation fields. This would impact the trained model’s efficacy and real-life applications. Based on that, they built the PlantDoc dataset, which can be a sufficiently large-scale non-lab dataset for leaf disease classification. The images in PlantDoc have cluttered backgrounds and are without a standard format. A comparison between Plant Village images and PlantDoc images can be seen in Figure \ref{fig:Apple_Scab_Leaf}. PlantDoc has similar categories to Plant Village with 2,598 leaf images from 13 plant species. In this dataset, there are 17 unique disease categories and 38 classes for the combination of species and diseases (e.g., Apple Scab Leaf). The images were annotated by experts.

 \vskip 1cm
\noindent\rule{4cm}{0.8pt}
\textbf{Take-home Messages}\noindent\rule{4cm}{0.8pt}
\begin{itemize}
    \item [1.] Maize Leaf (NLB) dataset is the largest public dataset while Plant Village is the most popular dataset. 
    
    \item [2.] Plant Village, Plant Leaves and Plantae K  are laboratory datasets which can be useful for prototyping and evaluating machine learning models. However, real-field datasets would provide a more comprehensive evaluation and support for realistic applications.
    \item [3.] We found that the available datasets are very useful for domain-adaptation and multi-task learning, however, this is largely missing in the current literature. We would suggest a machine learning model to learn from different datasets in a compositional manner where the model can effectively adapt to new tasks/datasets added in.
\end{itemize}
\noindent\rule{12cm}{0.8pt}
 \section{Machine Learning Approaches}

\label{sec:Approaches}
\begin{figure}[h!]
	\centering
	\includegraphics[width=1\textwidth]{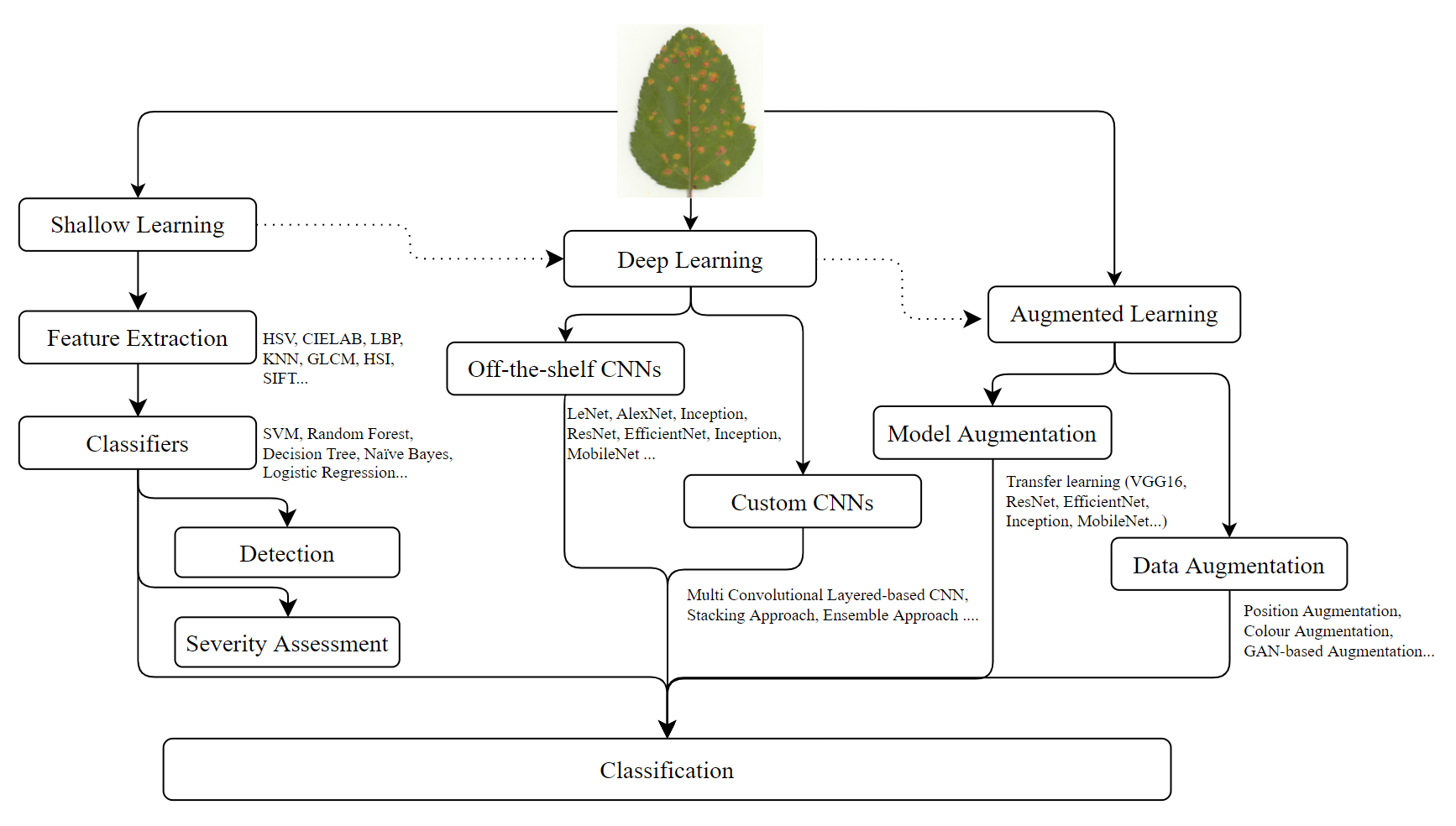}
	\caption{ML Development in Leaf Disease Classification}
	\label{fig:ML_Development}
\end{figure}

Generally, there are currently three general directions for machine learning approaches for leaf disease classification (see Figure \ref{fig:ML_Development}), including shallow learning (SL), deep learning (DL), and augmented learning (AL). In shallow learning approaches, leaf localisation always was done first, then based on the diseased leaves to classify the diseases. In addition, feature extraction is the necessary step of shallow learning to extract the features of leaves before classification. Deep learning has been emerging as a great tool for leaf disease classification recently thanks to its ability to offer an end-to-end process for learning and prediction. Deep learning does not require the feature engineering step and is able to learn an effective classifier from input images. At present, the advantages and disadvantages of shallow learning and deep learning approaches are still inconclusive. However, there is a strong agreement that SL has disadvantages in leaf image classification tasks, such as the inability to apply to large datasets, complex processing pipelines, and especially the need for feature extraction \cite{9388488, 9399342}. DL, however, also has two main disadvantages: computationally expensive and data-hungry. With the development of related hardware and computing systems, the computation expensiveness of DL has been alleviated. For the data hungriness issue, recent approaches employ augmented learning techniques by generating artificial data and/or reusing pre-trained models from other domains/tasks.

\begin{table}[h!]
	\begin{center}
		\begin{minipage}{\textwidth}
			\caption{Machine Learning Technologies}\label{Machine_Learning_Technologies}%
			\resizebox{\textwidth}{!}{%
				\begin{tabular}{@{}lcp{0.15\textwidth}cp{0.15\textwidth}p{0.30\textwidth}p{0.35\textwidth}@{}}
						\toprule
						Paper & Year & Dataset & Categories& Size (training/test) &Feature Extraction & Accuracy \\
						\midrule
						\cite{9362812} & 2020 & Plant Village (Part) & 2 & 250/150 & K-means & SVM (94.1\%)  \n
						\cite{9130019} & 2020 & Plant Village (Grape)& 4 & 70\%/30\% & LBP & SVM (97.76\%) \n			
						\cite{9077134} & 2020 & Plant Village (Part) & 2 & 126/54 & ROI & SVM (97.2\%) \n
						\cite{9182128} & 2020 & Plant Village (Tomato) & 8 & 60\%/40\% & GB & LR(67.3\%), RF(70.05\%) \& SVM (87.6\%)  \n	
						\cite{9210294} & 2020 & Plant Village (Tomato)  & 10 & 70\%/30\% & Colour Mean Pixel Value, Colour Moments, Prewitt operator, Gabor, Histogram Features, Haar, Histogram of Oriented Gradients (HOG) \& Local Binary Patterns (LBP) & HOG (the best FE) with, LR (Acc: 46.93\%, F1: 44.17\%), KNN (Acc: 44.10\%, F1: 44.10\%), \textbf{SVM (Acc: 48.77\%, F1: 40.38\%)}, Naïve Bayes (Acc: 25.88\%, F1: 27.22\%) \& DT (Acc: 39.74\%, F1: 39.66\%)  \n			
						\cite{a_15100615120210401}  & 2021 & Plant Village (Tomato) & 5 & N/A & LBP \&SIFT & SIFT \& MLP (92.40\%), SIFT \& RF (91.20\%), LBP \& MLP (90.40\%) \& LBP \& RF (89.30\%) \n
						\cite{9422499} & 2021 & Rice Leaves (Self) & 2 & 420/80 & Color Histogram, Hu Moments \& GLCM & \textbf{RF (97.50\%)}, Naïve Bayes, DT, LR, KNN \& SVM \n	
						\cite{7746160}& 2016 & Grape Leaves (Self)  & 2 & 110/27 & K-means & SVM (88.89\%) \n	
						\cite{9076371}& 2020 & Grape Leaves (Self)  & 2 & 60/30 & GLCM & SVM (90\%) \& \textbf{KNN (96.66\%)}  \n	
						\cite{8944556} & 2019 & Leaves (Self) & 2 & 55/20 & GLCM & \textbf{KNN (98.56\%)} \& Linear SVM (97.6\%) \n	
						\cite{9418680} & 2021 & Citrus Leaves (Self) & 4 & 170/170 & GLCM & SVM (91.76\%) \n
						\cite{9277379} & 2020 & Leaves (Self)  & 5 & N/A & K-means \& GLCM & SVM (Around 96\%) \n
						\cite{9212816} & 2020 & Banana Leaves (Self) & 4  & 371/247 & K-means &  SVM (85\%) \n
						\cite{edssjs.30A9493B20210101} & 2021 & Tea Leaves (Self) & 6 & 312 (4-fold) & PCA & SVM (83\%) \n	
						\cite{9142988} & 2020 & Leaves (Self) & 5 & N/A & N/A & Decision Tree (96\%) \\	
						\bottomrule
					\end{tabular} %
				}
			\end{minipage}
		\end{center}
	\end{table}

\subsection{Shallow  Learning} 

Table \ref{Machine_Learning_Technologies} summarises the details of this study through shallow machine learning approaches. We focus on the recent and notable papers from 2019. The general stages for leaf disease identification and classifications using shallow learning include: data(image) acquisition, processing, segmentation (possibly \cite{9725870, 9587775}), feature extraction, and identification (or classification) \cite{9396023, 9399342, 9725870, 9587775}. While data acquisition, processing, and segmentation are common in image processing generally, in this section we discuss two aspects that directly affect the quality of leaf disease classification.
\subsubsection{Feature Engineering}
Normally, data was collected from digital cameras (sometimes specialised cameras are used) to obtain basic features in colour models, such as RGB \cite{9212816}, HSV \cite{9362812,edssjs.30A9493B20210101}, and CIELAB \cite{9212816,9362812}.  Among the three colour models, HSV is more popular than the others. For example, \cite{9212816} collected 618 images from farms in RGB format before being converted to CIELAB colour space and resized to 400*600 pixels. In \cite{9362812}, the authors used two colour models (HSV and CIELAB) for Plant Village data to perform the segmentation for feature extraction.  \cite{edssjs.30A9493B20210101} collected 312 samples of tea leaves from three Indian tea gardens and convert them from RGB format to HSV for data pre-processing. From a colour model, we can extract more task-related features based on the spatial structure of the image data. The two most common methods for feature extraction are K-means clustering \cite{9362812,7746160,9277379,9212816} and grey-level co-occurrence matrix (GLCM) \cite{9076371,8944556,9418680,9277379,9422499}. From the literature, we found that GLCM features achieves better performance than K-Means features. Other extraction methods from image processing are employed as well. In \cite{edssjs.30A9493B20210101}, the authors used Non-dominated Sorting Genetic Algorithm (NSGA-II) to detect the tea leaf’s disease area and then applied Principal Component Analysis (PCA) to extract 5 most significant features for classification. In \cite{9077134} features are extracted from RoI (Region of Interest) segmentation. In \cite{9182128}, Gaussian blur and Haralick’s algorithm are applied to extract 60  texture features. A comparison of different feature extractors was presented in \cite{9210294}. In this study, 9 different feature extraction methods are used, including  Colour Mean Pixel Value, Colour moments, Edge Feature extraction using the Pewit operator, Gabor features extraction, Histogram features extraction, Haar features, Histogram of Oriented Gradients (HOG), and Local Binary Patterns (LBP). Among them, HOG features perform the best. Besides standard approaches in image processing, a novel feature extraction method based on Local Binary Patterns (LBP), dedicated to leaf diseases, was proposed in \cite{9130019}. This paper claimed that compared to recent grayscale LBP-based approaches, the new feature extraction method improved accuracy, precision and recall significantly.

Combination of features extracted from different techniques. \cite{a_15100615120210401} pre-processed all tomato leaf images through the Gaussian filtering (GF) technique first. After that, they tried to combine two feature extractors which are local binary patterns (LBP) and Scale Invariant Feature Transform (SIFT)

\subsubsection{Classifiers}
SVM was the most common ML classifier to classify the leaf diseases \cite{9362812,9130019,9077134,9182128,9210294,9422499,7746160,9076371,9418680,9277379,9212816,edssjs.30A9493B20210101}. \cite{7746160} used Linear SVM to detect the grape leaf disease, achieving 88.98\% accuracy. However, the linear kernel only works well if the data is linearly separated, which is not the case in many applications. In \cite{9362812}, a study compared three different kernels of SVM (Linear, Polynomial, RBF) on HSV and CIELAB features for Black rot disease classification in grape plan. The result showed that a SVM model with RBF Kernel gained the best accuracy of 94.1\%.  SVM was reported to be applied successfully to Banana leaf (85\% average accuracy) \cite{9212816}, tea leaves \cite{edssjs.30A9493B20210101} (83\% average accuracy, 78\% F1-score), grape vine disease (97.2\% average accuracy) \cite{9077134}. A comparison between SVM and Logistic Regression has been studied in  \cite{9182128} for tomato leave disease classification. The results showed that SVM significantly outperforms Logistic Regression (20\% better accuracy) and Random Forest (17\% better accuracy). In \cite{9210294} a more comprehensive comparison has been carried out with 4 competitors (Linear Regression, KNN, SVM, Naïve Bayes and Decision Tree) using 9 different types of features. It also concluded that SVM performs the best on tomato leaf disease diagnosis and severity measurement. A new SVM model was proposed in \cite{9418680}, known as hierarchical SVM, to detect citrus leaf diseases where hierarchical SVM achieved 91.76\% accuracy in comparison to 88.24\% from traditional SVMs.

Besides SVM, other classifiers can achieve high performance if suitable features are selected. For example, in a small private dataset, the performance of K-Nearest Neighbor (KNN) is 98.56\%, which is better than 97.6\% from SVM  \cite{8944556}. In \cite{9076371}, KNN  outperforms SVM when using GLCM features for grape leaf images, achieving  96.66\% in comparison to 90\% from the latter. For rice leaf disease
classification  \cite{9422499}, six ML algorithms, including RF, Naïve Bayes, Decision Trees, Logistic Regression, KNN and SVM, are compared. The feature set is a combination of Color Histogram, Hu Moments shape features, and Haralick texture features, which enabled RF to achieve the best performance (97.50\% accuracy) on an IoT device (Raspberry Pi). \cite{a_15100615120210401} pre-processed all tomato leaf images through the Gaussian filtering (GF) technique firstly. After that, they tried to combine two feature extractors which are local binary patterns (LBP) and Scale Invariant Feature Transform (SIFT) with two ML classifiers which are multilayer perceptron (MLP) and random forest (RF) models to classifier the tomato diseases. They measured the accuracy results of each feature extractor with each classifier, which are SIFT \& MLP 92.40\%, SIFT \& RF 91.20\%, LBP \& MLP 90.40\% and LBP \& RF 89.30\%. Decision Tree is a simple classifier and can be useful for small datasets with a small number of classes \cite{9142988}. Here, the paper shows that after relabelling the classes from four diseases and 1 healthy label to be a binary class, containing ‘healthy’ and ‘unhealthy’ labels, Decision Tree can achieve 96\% accuracy.

\vskip 1cm
\noindent\rule{4cm}{0.8pt}
\textbf{Take-home Messages}\noindent\rule{4cm}{0.8pt}
\begin{itemize}
	\item [1.] Shallow machine learning requires feature extraction from images\cite{9388488} to be useful for the disease classification task. The two most common methods are  K-means clustering and grey-level co-occurrence matrix (GLCM), in which GLCM is more recommended. A combination of features is also encouraged, as it can help improve performance.
	\item [2.] Support vector machine (SVM) was the most common ML method for leaf disease classification. It is very suitable for both smaller (more likely to be linear) or non-linear datasets\cite{9261801}. Its better performance in comparison to other classifiers is evident in several studies. However, if suitable features are selected, KNN or RF can achieve better accuracy.
	\item [3.] For small datasets with a small set of disease classes, simple methods can achieve good results.
\end{itemize}
\noindent\rule{12cm}{0.8pt}
\subsection{Deep Learning}
Deep learning is a rising branch of machine learning which consists of different architectures and associated learning algorithms. For leaf disease classification, most deep learning models and algorithms are based on neural networks with many number of hidden layers. We categorise deep learning approaches for this task into deep neural networks, convolutional neural networks for image classification, and convolutional neural networks for object detection\& classification. Table \ref{DL_Data_Augmentation} provides a summary of recent Deep Learning approaches for leaf disease classification.

\begin{table}[h!]
\begin{center}
\begin{minipage}{\textwidth}
	\caption{Summary of deep learning approaches.}\label{DL_Data_Augmentation}%
	\resizebox{\textwidth}{!}{%
		\begin{tabular}{@{}lcp{0.15\textwidth}cp{0.15\textwidth}cp{0.2\textwidth}p{0.4\textwidth}@{}}
				\toprule
				Paper & Year & Dataset & Categories& Size (training/test) & Augmentation & Features & Accuracy  \\
				\midrule
				\cite{9291694} & 2020 & Grape Leaves (Self) & 5 & 80\%/20\%  & Yes & Ensemble Method & Vanilla CNN (98\%), Improved VGG-16 (99\%), Improved MobileNet (97\%), Improved AlexNet (97\%) \&  \textbf{Ensemble (100\%)} \n
				\cite{10.3389/fpls.2020.00751}& 2020 & Grape Leaves (Self) & 4 & 4,449 images  & No & N/A  &  Faster R-CNN (81.1\%)  \n
				\cite{S187705092030690620200101} & 2020 & Plant Village (Tomato) & 10 &10,000/ 7,000/ 500\footnotemark[1] &  Yes & Assessed Storage Space & \textbf{CNN (91.2\%)}, Mobilenet (63.75\%), VGG-16 (77.2\%)\& InceptionV3 (63.4\%) \n
				\cite{Singh_2020} & 2020 & PlantDoc (Cropped) & 28 & 80\%/20\% &  Yes   & Created PlantDoc Dataset & VGG-16 (60.41\%), InceptionV3 (62.06\%) \& \textbf{InceptionResNet V2 (70.53\%)}    \n
				\cite{9261801} & 2020 & Grape Leaves (Self) & 6 & 80\%/20\% &   Yes &  Global Average Pooling (GAP) & VGG-16 with GAP (98.4\%) \n	
				\cite{8566635} & 2018 & Plant Village (Tomato) & 5 & 500(80\%/20\%)   & No  & with Learning Vector Quantization & CNN (86\%) \n	
				\cite{9350413} & 2020 & Tea Leaves (Self) & 3 & 1000/270/30\footnotemark[1] &  Yes  &  N/A & CNN (95.93\%) \n
				\cite{9408806} & 2021 & Plant Village (Modified) & 61 &31718/4540  &   Yes &  Stacking Method & \textbf{Stacking Model (87\%)}, ResNet (82.78\%), InceptionNet (82.22\%), DenseNet (83.44\%)\& InceptionResNet (84.07\%) \n	
				\cite{9250911} & 2020 & Betelvine Leaves (Self) & 3 & 1,014 images &  No  & Proposed Mask-RCNN (ResNet50 \& Feature Pyramid Network)  & F1: \textbf{Proposed Mask-RCNN (84.07\%)}, Faster RCNN (74.32\%) \& Mask RCNN (83.11\%) \n
				\cite{9250885} & 2020 & Plant Village (Whole) & 38 & N/A &  Yes &  Hybrid Approach: AlexNet + Linear SVM & \textbf{Hybrid Model (99.98\%)}, Basic AlexNet (96.34\%) \& AlexNet with GAP Layer (97.29\%) \n
				\cite{15107730820210601} & 2021 & Plant Village (Tomato) & 2, 6, 10 & 5-fold  &  Yes  &  N/A & EfficientNet B0, B4, B7 (97\% - 99\%) \n	
				\cite{9392051} & 2020 & Cassava Leaves\cite{mwebaze2019icassava} & 5 & 5,656/ 1,889/ 1,885\footnotemark[1] &   No & N/A  & MobileNet (85.38\%) \n	
				\cite{9342729} & 2020 & Plant Village (Peach, Pepper \& Strawberry) & 6 & 70\%/30\% &  N/A &  Tested different epochs (50, 75,100 \& 125) & Multi Convolutional Layered-based CNN (87.47\% - 99.25\%) \n	
				\cite{15100606020210401} & 2021 & Pepper Leaves (Self) & 2 & N/A &  N/A  &  Used GLCM as Feature Extraction  & \textbf{DBN (91.956\%\&77.546\%)}, FFNN (91.156\%\&63.936\%), BPNN (91.306\%\&66.916\%), DNN (91.386\%\&67.246\%), RNN (91.436\%\&67.486\%) \& CNN (91.616\%\&72.046\%) \footnotemark[2] \n	
				\cite{S004579062100047120210301} & 2021 & Tea Leaves (Self) & 3 & 318/80, 1400/200  & Yes & Diseased Leaf classification \& Disease Severity Analysis & Faster Region-based CNN (91.22\%) \n	
				\cite{t_14770449620201201} & 2020 & Plant Village (Tomato) & 10 & 80\%/20\% &  No & N/A & \textbf{Xception V4 (99.45\%)}, AlexNet (90.1\%), Lenet (88.3\%),  Resnet (98.40\%) \& VGG-16 (90.1\%) \n	
				\cite{8374024} & 2018 & Maize Leaves (Self \& Plant Village) & 9 & 80\%/20\% & Yes& N/A &  Fine-tuned GoogLeNet (98.9\%) \& Cifar10 (98.8\%)   \\
				%
				\bottomrule
			\end{tabular} %
		}
		\footnotetext[1]{Training/Validation/Test Amount}
		\footnotetext[2]{Acc \& F1}
		
	\end{minipage}
\end{center}
\end{table}

\subsubsection{Deep Neural Nets}
Deep neural networks are neural networks with multiple hidden layers, one on top of another. Previously, training such deep structures is difficult due to the problem of gradient vanishing/exploding but current learning techniques can turn that cure into a blessing, thanks to the availability of big data and powerful computing resources. We can use deep neural nets as a classifier, similar to shallow learning approaches. In \cite{15100606020210401}, deep Belief networks (DBN) were studied, together with other variants of multi-layer feedforward neural networks, for pepper leaf disease classification. The models were evaluated on two datasets. The first dataset is self-collected, consisting of 1500 images of healthy and diseased leaves. The other dataset contains 300 healthy and 35 diseased images from Plant Village. All samples are resized to 256 *256 pixels. The features used in this study was Gray Level Co-occurrence Matrix (GLCM). The average accuracy and F1-score of DBN are 91.956\% and 0.77546, respectively. The results are slightly better performance.

The employment of feature engineering in deep learning seems not useful, as deep models themselves are effective feature extractors. Instead of two stages (feature extraction + classification) deep convolutional neural networks (CNN) can learn discriminative features that are useful for classification in an end-to-end manner.

\subsubsection{Image Classification CNNs}
CNN is a class of neural networks where spatial information from image structure are represented and learned through convolution operations. CNNs have been used largely in image processing and computer vision, especially in classifying images, and therefore have been useful for leaf disease classification as well.

\noindent\textbf{Off-the-self CNNs}
There are a plethora of convolutional neural networks developed to tackle a wide range of problems in image classification. Ones can easily pick up a model and apply it to classify disease from leaf images.

{\it LeNet \& GoogLeNet} LeNet \cite{726791}
is one of the earliest convolution CNNs, although it does not have a very deep architecture, its convolution idea is the inspiration for many other deep CNNs models nowadays. In \cite{9250885}, LeNet achieved the lowest accuracy (94.0\%) compared to other approaches on Plant Village. A newer version, called GoogLeNet (also known as Inception V1), was developed with improvements from LeNet with several novel components added, such as batch normalization, image distortions, and more layers. In \cite{9418245} GoogLeNet achieved  95.69\% accuracy and ranked 3rd in 7 CNN models for Apple disease classification. In \cite{8374024} it achieved 98.9\% accuracy for the classification of Maize leaf diseases.

{\it AlexNet}. As one of the earliest deep CNN models, AlexNet has been employed in multiple studies of leaf disease classification \cite{S004579061930002320190601,9137986,8974752,t_14770449620201201,9291694,9250885}. For tomato diseases, AlexNet achieved promising results, such as 95.75\% accuracy in \cite{9137986} and 90.1\% accuracy \cite{t_14770449620201201} (they used different testing partitions).  AlexNet was also reported to have 86.5\% accuracy for grape diseases in \cite{8974752}. Although AlexNet was a popular model, its performance was usually inferior compared to other deep CNNs. For improvement, \cite{9250885} proposed a hybrid approach by combining  AlexNet and Linear SVM to boost the accuracy to  99.98\% on the Plant Village dataset. This is significantly better than AlexNet alone (94.3\%), ResNet50 (98.06\%), VGG-16 (98.76\%), and Inception V3 (99.08\%).

{\it VGG}. Very Deep Convolutional Networks, known as VGG or VGGNet, is an idea of how to effectively increase the depth of CNNs.  VGG-16 (VGG with 16 layers) has been applied to tomato leaves datasets \cite{S187705092030690620200101,t_14770449620201201}.
In  \cite{S187705092030690620200101} a pre-trained model was used to achieve $77.2\%$. In \cite{t_14770449620201201} a better training approach was proposed where the performance was much higher with 90.1\% accuracy. A deeper version of VGG, VGG-19, was employed in \cite{9231174} to successfully classify tomato leaf diseases with 96.86\% accuracy. In \cite{S004579062100047120210301} used VGG-16 to do the severity analysis The proposed model gained 91.22\% Accuracy. \cite{SUJATHA2021103615} applied VGG-16 and VGG-19 on a citrus leaf disease dataset. Notably, VGG-16 has been applied widely to grape leaf images \cite{8974752,9291694, 9261801}.  \cite{9261801} tested VGG-16 on their private grape leaf diseases dataset (5 leaf diseases and 1 healthy category,6000 images). Some modifications of VGG16 have been developed by replacing two last two fully connected layers with the Global Average Pooling layer. The results showed that the proposed has the best accuracy (98.4\%), significantly better than normal VGG-16 and the combination of VGG-16 and SVM classifier.

{\it Inception} Inception is a class of CNNs that utilises Inception modules for deeper structure with more efficient computation.  In leaf disease classification, Inception V3 was the most popular among different versions of Inception networks. It was employed for tomato leaf diseases  \cite{S187705092030690620200101}. In \cite{15100606920210401} Inception V3 achieved 95.41\%  on a rice diseases image dataset, better than VGG-16 and RestNet-50. For the benchmark Plant Village dataset, InceptionV3 was reported to receive 98.42\% \cite{electronics10121388}, and \textbf{99.74\%}, \cite{14844519620210101}. Again, they have different results because of the different partitions for training, validation, and test.

{\it ResNet} Among many deep CNN models, ResNet is a powerful structure where we can train the model with a lot of layers to gain performance superiority. ResNet-50 achieved   98.40\% accuracy for tomato leaves \cite{t_14770449620201201}. \cite{9408806} applied  ResNet to achieve 82.78\% in modified Plant Village. For Betelvine leaf disease,  \cite{9155585} showed that ResNet-34 outperformed other models with 99.40\% accuracy \&  0.9651 F1-score. These are much better than SVM (50.69\% \& 50.57\%), Decision Tree (72.23\% \& 72.02\%), Logistic Regression (80.99\% \& 80.88\%) and K-NN (87.86\% \& 88.06\%). Another version, ResNet-20, achieved 92.76\% on apple leaf images \cite{9418245}. Recent works integrate the idea of residual blocks in ResNet and Inception module \cite{electronics10121388} to create InceptionResNetV2. Such a combination increases the performance from 98.42\% to 99.11\% on the Plant Village dataset.

{\it MobileNet \& EfficientNet}. Besides very deep models as we discussed above, some compact architectures were also employed, thanks to the increasing demand for IoT and hardware devices in plant pathology. For example,  MobileNet can predict grape leaf diseases with 86\% accuracy \cite{9291694}. 
In \cite{9392051}, MobileNet was applied to predict diseases from Cassava leaves \cite{mwebaze2019icassava}. This public dataset has 1 healthy and 5 disease classes and was split into a training set (5,656 images), a validation set (1,889 images) and a test set (1,885 images). All images are resized to 224 * 224 pixels. The proposed MobileNet model gained 85.38\% accuracy. In \cite{S187705092030690620200101} MobileNet was shown to achieve 63.75\% on tomato leaf images. In \cite{15107730820210601}, the authors employed three sub-models (B0, B4, B7) of EfficientNet to classify tomato leaf diseases (Plant Village’s 10 tomato categories). There are three types of this study’s classification tasks, binary classification (healthy or unhealthy), six-class classification (1 healthy and 9 diseased categories are categorized into 5 classes, i.e., bacterial, fungal, viral, mold, and mite disease) and ten-class classification (1 healthy and 9 diseased). All images were resized to 224 $\times$ 224 and data augmentation was applied. The evaluation was carried out with 5-fold cross-validation. The results showed that for binary classification and six-class classification, EfficientNet-B7 had the best performance with an accuracy of 99.95\% and 99.12\%, respectively. For the ten-class classification, EfficientNet-B4 performed better than other models with an accuracy of 99.89\%.

\noindent\textbf{Custom CNN.} Although off-the-shelf CNN models were shown to be useful for leaf disease classification, they were originally designed and tested for general image classification tasks using benchmarking datasets, much different from leaf images. Therefore, they may not be optimal for this specific task and custom CNN models can be best for each dataset. Many researchers customised and developed their own CNN models, either from scratch or modify from existing ones.  In \cite{S187705092030690620200101}, a new CNN model was developed to classify tomato leaf diseases (extracted from Plant Village). They compared the proposed CNN model with Mobilenet, VGG-16 and InceptionV3. The proposed model’s accuracy is 91.2\%, better than the others, and its storage space is the smallest (1,696 KB). \cite{8566635} also studied tomato leaves from Plant Village. They used the CNN model with Learning Vector Quantization (LVQ) algorithm to classify the diseases. The model achieved 86\% average accuracy. Another variant of CNNs was proposed in \cite{9350413} to classify two tea leaf diseases. The precision of this model was approximately 95.93\%. In \cite{9342729}, the authors designed a new Multi Convolutional Layered-based CNN model and apply it to three sub-datasets (Peach, Pepper, and Strawberry) from Plant Village. They showed that their CNN can effectively classify the leaves of three sub-datasets with accuracy from 87.47\% to 99.25\%. The CNN model in \cite{t_14770449620201201} was developed based on Xception V4 architecture and was tested to compare with several common pre-trained models, including VGG-16, ResNet-50, AlexNet and LeNet. The dataset used in this study is 10 classes of tomato leaves from Plant Village, where 14528 images were split into 80\% for training and 20\% for testing.  The experiment results (in accuracy score) are: the proposed model (99.45\%), AlexNet (90.1\%), Lenet (88.3\%),  Resnet (98.40\%) and VGG-16 (90.1\%). \cite{9291694}  tested Vanilla CNN  and three pre-trained models (VGG-16, MobileNet \& AlexNet). Finally, they built an ensemble model (average voting method) which achieve perfect accuracy of 100\%.

A stacking approach was developed in \cite{9408806}, aiming to create an effective way to improve classification accuracy. The dataset in this work is from AI-Challenger 2018 (which was modified from Plant Village), it contains 10 different plant species and 61 classes. They split the dataset into a training set (31718 images) and a test set (4540 images). After data augmentation, the training set has been trained by four models (Inception Network, ResNet, Inception Combine ResNet and DenseNet), and being stacked. The stacking method achieved 87\% accuracy, better than ResNet (82.78\%), Inception Net (82.22\%), DenseNet (83.44\%) and Inception-ResNet (84.07\%).

Another idea is to employ a hybrid approach, between deep learning and shallow learning, where deep learning would play a role of a feature extractor \cite{9250885}. In this work, AlexNet was combined with Linear SVM to classify diseases in the Plant Village dataset (resized to 227 $\times$ 227 pixels). The experimental results showed that their proposed model gained  99.98\% accuracy better than the basic AlexNet (96.34\%) and AlexNet with Global Average Pooling Layer (97.29\%). In addition, they evaluated different optimizers (AdaMax, AdaDelta, Adam, RMS Prop, SGD, AdaGrad)  and showed that AdaMax has the best performance in this study.

\subsubsection{Object Detection \& Classification CNNs}
In real-life scenarios, it would be useful if a system can detect leaves from cluttered backgrounds and classify their diseases. In this case, image segmentation can be applied as a first stage to extract the leaves area before applying CNNs for image classification as we discussed in the previous section. However, it would be more convenient to have an end-to-end approach where CNNs can detect leaves and identify diseases. In \cite{10.3389/fpls.2020.00751}, the authors employed a
Faster Region-based CNN (R-CNN) model to detect and classify grape leaf disease with the best accuracy of 81.1\%. Faster R-CNN was also the interesting model in \cite{Singh_2020} for an evaluation of the PlantDoc dataset. They claimed that fine-tuning Faster R-CNN with InceptionResnetV2 and MobileNet can reduce the classification error significantly.  \cite{S004579062100047120210301} proposed a model based on Faster R-CNN to detect tea leaf blight (TLB) and used VGG-16 to do the severity analysis. The dataset of disease classification has 398 images. Among them, 80 made up the test set. The dataset of severity analysis contains 270 mildly diseased leaf images (after augmentation, it increased to 700) in the training set and 100 in the test set, 700 Severe diseased leaf images in the training set and 100 in the test set. The proposed model gained 91.22\% accuracy. \cite{9250911} studied another variant of R-CNN, namely Mask R-CNN. They improved Masked-RCNN with ResNet50 and Feature Pyramid Network as key components, to classify Betelvine leaf diseases. For evaluation, a private dataset was collected from real cultivated Betelvine crops containing two diseases which are Anthracnose (358 images) and Phytophthora (456 images), and 1 healthy category (200 images). All images are resized to 256 * 256 pixels. The proposed Mask-RCNN model achieved 84.07\% F1-score, which is better than Faster-RCNN (74.32\%) and the original Mask-RCNN (83.11\%).

\subsubsection{Comparison between DL and SL}
\begin{table}[h!]
	\begin{center}
		\begin{minipage}{\textwidth}
			\caption{Comparison Between ML \& DL}\label{DL_Comparison}%
			\resizebox{\textwidth}{!}{%
				\begin{tabular}{@{}lcp{0.15\textwidth}cp{0.15\textwidth}p{0.21\textwidth}p{0.4\textwidth}@{}}
						\toprule
						Paper & Year & Dataset & Categories& Size (training/test) &Feature Extraction &  Accuracy \n
						\cite{9057889} & 2020 & Plant Village (Whole)  & 19 & N/A & K-means & LR (66.4\%), KNN (54.5\%) \& SVM (53.4\%)  \\	
						& & & & & None & \textbf{CNN (98\%)} \n		
						\cite{9418013} & 2021 & Plant Village (Part) & 15 & 80\%/20\% & Co-occurrence Matrix & \textbf{CNN (96\%)}, ANN (90\%), KNN (88.6\%), SVM (85\%), Naïve Bayes (79.6\%) \& K-means (72.3\%)  \n
						\cite{15142389920210101} & 2021 & Plant Village (Tomato) & N/A & N/A  & K-means \&GLCM &  CNN (N/A) \n
						\cite{8974752} & 2019 & Plant Village (Grape) & 4 & 3800/200 &  HSV-histogram & Decision Tree (80.5\%), Naive Bayes (69.5\%), SVM (85\%), LDA (80.5\%), KNN (96\%), LR (94\%), RF (97.5\%), ANN (87.5\%) \\				 
						&  &    &   &   &  None &  \textbf{CNN (99\%)}, AlexNet (86.5\%), VGG-16 (97.5\%) \n							
						\cite{S004579061930002320190601} & 2019 & Plant Village (Whole)  & 38 & 55,636/1950 & N/A & SVM (50.69\%), Decision Tree (72.24\%), LR (81.00\%) \& K-NN (87.87\%) \\
						&  &  & &  & None &\textbf{CNN (97.87\%)}, AlexNet (87.34\%), ResNet (92.56\%), VGG-16 (92.87\%) \& Inception-v3 (94.32\%) \n	
						\cite{9155585} & 2020 & Plant Village (Modified) & 38 & 15,200 (80\%/20\%) & GLCM & SVM (50.69\% \& 50.57\%), DT (72.23\% \& 72.02\%), LR (80.99\% \& 80.88\%) \& K-NN (87.86\% \& 88.06\%) \\	
						&  &  & &  & None & \textbf{ResNet34 (99.40\% \& 96.51\%)}\footnotemark[1]  \n						
						\cite{9418245} & 2021 & Plant Village (Apple) & 4 & 10888/2801 & N/A &  SVM (68.73\%) \& BP (54.63\%) \\
						& & & & & N/A & \textbf{CNN (97.62\%)}, AlexNet (91.19\%), GoogLeNet (95.69\%), ResNet-20 (92.76\%) \& VGG-16 (96.32\%)  \n	  
						\cite{9137986} & 2020 & Tomato Leaves (Self) & 4 & N/A  & DWT \& GLCM &  \textbf{CNN (98.12\%)}, AlexNet (95.75\%) \& ANN (92.94\%)  \n
						\cite{SUJATHA2021103615} & 2021 & Citrus Leaves\cite{RAUF2019104340} & 5 & 10-fold & N/A &  RF (76.8\%), SGD (86.5\%) \& SVM (87\%) \\
						& & & & & None & VGG-19 (87.4\%), Inception-v3 (89\%) \& \textbf{VGG-16 (89.5\%)}   \\
 					
						\bottomrule
					\end{tabular} %
				}
						\footnotetext[1]{Acc \& F1}
				
			\end{minipage}
		\end{center}
	\end{table}

Early applications of deep learning attempted to integrate deep models with feature extraction. For example, in  \cite{15142389920210101} and \cite{9137986}, the authors employed hand-crafted features for image segmentation before training CNNs to classify the tomato leaf diseases. In particular, \cite{15142389920210101} 
 employed k-means clustering for feature extraction, coupled with CNNs to estimate disease severity, although their results are not clearly detailed. In \cite{9137986} Discrete Wavelet Transform (DWT) and grey-level co-occurrence matrix (GLCM) features were used to segment leaves from the background which helped a CNN model to achieve  98.12\% accuracy, better than AlexNet (95.75\%) and traditional (shallow) neural networks (92.94\%).
 
Comparisons between SL and DL methods have been carried out largely in recent years. When applying them on the same datasets the performance of DL methods tends to be superior. Deep learning approaches, such as CNNs, are very effective in image classification where abundant data is available as CNNs can extract discriminative features from images automatically. Therefore, the descriptiveness of feature extractors used in shallow learning can be a bottleneck for classifying leave diseases from images. We show the details of the current comparison in Table \ref{DL_Comparison}.
\cite{SUJATHA2021103615} compared the performance of SVM, RF, Stochastic Gradient Descent (SGD), Inception-V3, VGG-16 and VGG-19 on the citrus leaf disease dataset. Using 10-fold cross-validation, 3 deep learning methods were shown better than the shallow counterpart. A study in \cite{9057889} compared logistic regression(LR), KNN, and SVM with CNN on the Plant village dataset. The shallow learning methods in this work used K-means clustering as the feature extractor. The experimental results demonstrated that CNN got an overwhelming victory (98\% accuracy) compared to other ML methods (around 60\%). A deeper study has been shown in \cite{9418013} where the authors analysed the weaknesses of several shallow learning methods, including K-Means, (shallow) artificial neural networks (ANN), Naïve Bayes, SVM, and KNN. For the empirical results, K-Means and ANN have quite low accuracy, and Naïve Bayes has a slow convergence rate. Meanwhile, SVM achieves relatively poor performance and KNN has some dimensionality issues. The such analysis led to an investigation into a system based on CNNs to improve the performance. As expected, the proposed CNN achieved the best accuracy (96\%). \cite{8974752} used general data augmentation methods i.e. zooming, inversion, flipping, rotation, to make the training free from bias for any particular class (a.k.a balancing data). In this work, the CNN model also achieved the best accuracy of 99\%. This is better than other pre-trained models they tested (AlexNet: 86.5\%, VGG-16: 97.5\%), and also other shallow learning approaches (Decision Tree, Naive Bayes, SVM, LDA, KNN, LR and RF). Among the shallow learning models, RF with HSV-histogram feature achieved the best result (97.5\%). The proposed CNN model in \cite{S004579061930002320190601} can classify leave diseases with 97.87\% accuracy, better than the popular transfer learning approaches (AlexNet, VGG-16, Inception-v3 and ResNet) and shallow learning approaches (SVM, logistic regression, decision tree and K-NN). In another work \cite{9155585}, the authors employed Residual Networks (ResNet34) to construct a custom model with 99.40\% accuracy and 96.51\% F1-score. This results significantly surpass shallow learning models: SVM (50.69\% \& 50.57\%), Decision Tree (72.23\% \& 72.02\%), Logistic Regression (80.99\% \& 80.88\%) and K-NN (87.86\% \& 88.06\%). \cite{9418245} used their proposed approach (integrating CNN with AlexNet and GoogLeNet cascade inception) to classify apple leaf diseases. Their proposed model gained 97.62\% better than shallow learning, including SVM (68.73\%) and Back Propagation (54.63\%).

From multiple studies on the comparison between shallow learning and deep learning, some researchers concluded that compared with the shallow learning approaches the deep learning approaches, based on CNN architecture, can be more suitable and effective for leaf disease classification \cite{9057889}. As we can see, CNNs do not require manual pre-processing or feature extraction which may cause side effects \cite{sharma2021detection, 9231174}, although it can shorten the training time and fewer computations for shallow learning. Table \ref{DL_Comparison} clearly shows that CNNs outperform shallow learning by a significant margin. However, if the data is small, shallow learning can be more useful \cite{8374024}. In order to make deep learning effective, the quantity of data should be sufficient. In the next section, we will show how augmentation has been emerging as a great tool to deal with the data availability problem.

\vskip 1cm

\noindent\rule{4cm}{0.8pt}
\textbf{Take-home Messages}\noindent\rule{4cm}{0.8pt}
\begin{itemize}
    \item Deep learning models are useful for leaf disease classification and should be recommended in real-life applications due to their high accuracy. The common off-the-shelf deep learning models are CNN, AlexNet, VGG-16, ResNet, EfficientNet, Inception and MobileNet. 
    \item Custom CNNs are highly encouraged as we should design an optimal model for different tasks. It was evident that custom CNNs perform better than off-the-shelf models.
    \item Deep learning is more effective than shallow learning in leaf disease classification. It is also more convenient as we can get rid of the feature extraction steps and minimise the manual effort for data processing.
    \item Compare with Table \ref{Machine_Learning_Technologies}, we can see that the datasets used in deep learning papers were relatively larger than in other studies. This is consistent with the fact that deep learning models are usually data-hungry.
    \item Most of the studies focus on the performance (accuracy) aspect of the task while a more comprehensive comparison with compactness and efficiency is still missing. There are a few papers that addressed these issues, for example, \cite{9057889} evaluates models' speed and  \cite{S187705092030690620200101} evaluates models' storage space.
    \item Different studies use different experiment settings, including different partitions for training/validation/test which makes their results difficult to compare. Therefore, a benchmarking study is needed.
\end{itemize}
\noindent\rule{12cm}{0.8pt}

\subsection{Augmented Learning}

\subsubsection{Data Augmentation}

As mentioned previously, the main obstacle to this research is the availability of datasets \cite{edssjs.15D0A2D220200101, 531183386, 9399342,9544640}. More often situations researchers need to deal with the problem of not having enough data (i.e., small datasets) first, therefore, data augmentation is an effective method to solve this problem. Data augmentation can be seen as the imagination or dreaming of humans where we can simulate different scenarios based on our experience to anticipate unobserved events \cite{shorten2019survey}.

Many research results have already confirmed the effectiveness of data augmentation in leaf disease classification \cite{S004579061930002320190601, 9238318, 524942744, 9752495, 9823799, exsy.12885}. Table \ref{Data_Aug} shows the common data augmentation methods in leaf disease classification. Data augmentation has several purposes, as follows: i) enrich a dataset by increasing its volume \cite{9408806,9331214}; ii) mitigate the data imbalance problem \cite{9399342, S004579062100047120210301}; iii) improve the generality to reduce the over-fitting issue and make machine learning models more robust \cite{9408806, 9331214, 9238318, exsy.12885}. Generally, in leaf disease classification the common data augmentation approaches (including physical expansion \cite{9399342} and position and colour augmentation \cite{524942744}), are widely used thanks to their convenience and simplicity. There are many existing functions and tools available for position augmentation, such as Pytorch’s transforms function (torchvision.transforms) \cite{9138030} and the Augmentor python library \cite{S187705092030690620200101}. Position augmentation methods mean changing the image’s position, shape, size and so on. Rotating (rotation) is the most used method, as can be seen in \cite{S004579061930002320190601, 8974752, 9291694, 8374024, 9331214, 9397001, 15107730820210601, 9418013, 9138030, S187705092030690620200101, JEPKOECH2021107142, 9231174, S004579062100047120210301, 524942744, 9752495, 9823799, 517091827}. Here, the method rotates leaf images to different angles (e.g. 30\degree, 90\degree  or 180\degree ) to produce new samples. After rotating, we can apply other techniques to generate more samples, such as flipping \cite{S004579061930002320190601, 8974752, 9291694, 9331214, 9397001, 9418013, 9138030, S187705092030690620200101, JEPKOECH2021107142, 9231174, 524942744, 9752495, 9823799, 517091827}, zooming/scaling \cite{8974752, 9291694, 9231174, 9823799, 517091827, S004579061930002320190601,  8374024, 15107730820210601, 9138030, 524942744}, cropping \cite{9397001, 9418013, 9418013, 9418013, S187705092030690620200101, 524942744}, vertical or horizontal shearing \cite{517091827, 9291694, 9418013}, shifting \cite{9291694, 9418013, 9231174,  9752495, 517091827}, transformation \cite{9397001, 9138030, S004579062100047120210301, 524942744}, translation \cite{9331214, 15107730820210601, 9138030, 524942744}; and resizing \cite{9138030, S187705092030690620200101}.

Besides texture augmentation, researchers also used colour augmentation to process the leaf images, such as Brightness, contrast, saturation, hue \cite{524942744, 9397001}, and Principal Component Analysis (PCA) colour augmentation \cite{S004579061930002320190601}. It is worth noting that there may be pitfalls to the use of colour augmentation techniques for leaf images as colour is important to identify diseases. Therefore, we should be careful not to destroy or alter the original features of the leaf images. For example, some researchers used colour augmentation methods to change colourful leaf images \cite{8374024, 524942744,S004579061930002320190601}, but in \cite{9399342} the authors pointed out that colour may be one of the most important manifestations of some leaf diseases, so changing the colour features of original images may bring negative effects.

The augmentation methods mentioned above may have limitations such as poor quality, inadequate diversity, and unevenness \cite{9399342}. Recent approaches, including Generative Adversarial Networks (GAN) \cite{NIPS2014_5ca3e9b1}, employ deep learning to generate artificial data. \JY{GAN techniques employ a neural networks called generator to produce images which are different from a training set  to fool a classifier (a discriminator) as if they belong to some classes of the set. In the case of leaf images, GAN can generate new images for different disease types}. Compared to the non-learning methods, GAN‑based Data Augmentation is based on generative modelling and learning where the focus is on creating artificial samples and retaining similar characteristics from the original dataset. GAN has been widely used to create more samples recently \cite{9399342}. In \cite{9823799}, \JY{the original dataset comprises a total of 3941 images, including 1858 images of bacterial blight and 1706 images of leaf blast. After applying GAN augmentation, the dataset size increased to 9101 images, with 3767 images representing bacterial blight and 5034 images representing leaf blast,} and the experimental results showed that the accuracy of CNN models can be improved with data generated from GAN.

Besides the texture/colour-based transformation and GAN approaches, there are some new methods were developed. For example, \cite{exsy.12885} proposed two image augmentation (IA) methods, including image pre-processing \& transformation algorithm (IPTA) and image masking \& REC-based hybrid segmentation algorithm (IMHSA). The methods aim to produce a sufficient quantity of training leaf disease images to improve the richness of small datasets. IPTA is an adaptive supervised learning approach to transform the original images into augmented images. IMHSA is an unsupervised approach for RGB image segmentation. The empirical study showed that with augmented data the validation accuracy was raised from 65\% to 73\%.

\begin{table}[h!]
	\begin{center}
		\begin{minipage}{\textwidth}
			\caption{Common Data Augmentation Technologies in Leaf Disease classification}\label{Data_Aug}%
			\resizebox{\textwidth}{!}{%
				\begin{tabular}{@{}lcp{0.15\textwidth}p{0.5\textwidth}p{0.25\textwidth}@{}}
						\toprule
						Paper & Year & Datasets  & Augmentation & Others   \\
						\midrule
						\cite{S004579061930002320190601}& 2019 & Plant Village (Whole)  & Image flipping, gamma correction, noise injection, PCA colour augmentation, rotation \& scaling &   Increase the performance  \n
						\cite{8974752} & 2019 &  Plant Village (Grape) & Zooming, inversion, flipping \& rotation &     \n		
						\cite{9291694} & 2020 & Grape Leaves (Self) & Rotating, shift, shear, zoom and horizontally flip &  \n
						\cite{8374024} & 2018 & Maize Leaves (Self \& Plant Village) & Rotating, cutting \& grayscale &  \n
						\cite{9331214} & 2021 & Leaves (Self) & Flipping, rotation, translation, channel drop & \n
						\cite{9397001} & 2021 & Plant Village (Tomato) & 	Cropping, flipping, rotating, transformation, noise rejection, colour augmentation &  \n
						\cite{8971580} & 2019  & Plant Village (Whole) & Neural Style Transfer (NST) and Generative Adversarial Networks (GANs)&  \n
						\cite{15107730820210601} & 2021 & Plant Village (Tomato) & Rotating, scaling, and translation &  \n
						\cite{9418013} &  2021 & Plant Village (Part) & 	Rotating, shifting, flipping, cropping, and shearing &  \n
						\cite{9138030}&   2020 & Plant Village (Tomato)   & Random Rotation, Random Resized Crop, and Random Rotation \& Resized Crop & torchvision.transforms (Pytorch) \n
						\cite{S187705092030690620200101} &  2020 & Plant Village (Tomato)   & Rotating, flipping, cropping and resizing  & Augmentor Package (python) \n
						\cite {JEPKOECH2021107142} & 2021 & JMuBEN \& JMuBEN2 & Rotating \& flipping &  \n
						\cite{9231174} &  2020 & Plant Village (Tomato)  & Rotating, vertical and horizontal flips, linear shifting and brightness \& zoom variation &  \n
						\cite{9399342}& 2021 & N/A & Generate Adversarial Networks (GANs)  &  \n
						\cite{S004579062100047120210301} & 2021 & Tea Leaves (Self)  & symmetrical transformation \& rotating &  \n
						\cite{524942744} & 2022 & N/A &\begin{tabular}{@{}p{0.5\textwidth}@{}} Position augmentation: Scaling, cropping, flipping, padding, rotation, translation \& affine transformation \\ Colour Augmentation: Brightness, contrast, saturation \& hue \end{tabular} &  \n
						\cite{9752495}& 2022 & Plant Village (Whole) & Image blur, sharpening, contrast enhancement, shifting, rotation and flipping & Accuracy was raised from 97.59\% to 99.5\% \n
						\cite{9823799} & 2022 & N/A & Horizontal flipping, rotating, zooming \& GANs augmentation & Increase images from 3941 to 9101 \n
						\cite{exsy.12885}  & 2021 & Plant Village (Whole)  & IPTA \& IMHSA & Accuracy was raised from 65\% to 73\% \n
						\cite{9753846} & 2022 & Plant Village (Apple \&Grape) & Picture editing, changing over \& upgrade &  \n
						\cite{517091827} & 2022 & Plant Village (Tomato)  & Image zooming, vertical shearing, horizontal shearing, vertical flip, horizontal flip, vertical shift,  horizontal shift \& rotating &  \\

%

						\bottomrule
					\end{tabular} %
				}

			\end{minipage}
			
		\end{center}
	\end{table}

\subsubsection{Model Augmentation (Transfer learning)}

\begin{table}[h!]
	\begin{center}
		\begin{minipage}{\textwidth}
			\caption{Transfer Learning}\label{Transfer_Learning}%
			\resizebox{\textwidth}{!}{%
				\begin{tabular}{@{}lcp{0.17\textwidth}ccp{0.5\textwidth}@{}}
						\toprule
						Paper & Year & Dataset &  Augmentation & Pre-train & Accuracy  \n
						\cite{9355991} & 2020 & Apple \& Grape Leaves (Self) &  No  & ImageNet & Fine-tuned VGG-16 (97.87\%) \n	
						\cite{9293866} & 2020 & Plant Village (Grape)  &  No &  N/A & \textbf{AlexNet (95.65\%)}, GoogLeNet (92.29\%) \& ResNet-18 (89.49\%) \n	
						\cite{9331214} &2021 & Leaves (Self)  &  Yes  & N/A  & Mobilnetv2 (90.38\%)  \n
						\cite{15100606920210401} & 2021 & Rice Diseases Image Dataset \footnotemark[1]&  Yes &  ImageNet & VGG-16 (87.08\%), ResNet50 (93.41\%) \& \textbf{InceptionV3 (95.41\%)} \n	
						\cite{9393311} & 2021 & Plant Village (Tomato) &  No & ImageNet  &  VGG-16 (98.00\%) \& \textbf{GoogLeNet (99.23\%)} \n	
						\cite{9278067}& 2020 & Plant Village (Tomato) &  No & ImageNet  & Base VGG-16 (63.53\%),  \textbf{Pre-trained VGG-16 (77.65\%)} \& VGG-16 with L2 Regularisation (72.94\%) \n	
						\cite{sharma2021detection} & 2021 & Plant Village (Part) &   No & ImageNet & F1-score: ResNet-50 with GAP \& DCT Compression  (92\%), \textbf{ResNet-50 with GAP (98\%)} \n	
						\cite{9138030} & 2020 & Plant Village (Tomato)  &  Yes  &  ImageNet  & ResNet-50 (97\%) \n	
						\cite{9231174} & 2020 & Plant Village (Tomato)  &  Yes & ImageNet & Segmented: CNN (83.46\%), MobileNetV2 (96\%), \textbf{EfficientNet-B0 (96.40\%)}, VGG-19 (92.06\%) ,  Unsegmented: CNN (81.46\%), MobileNetV2 (97.26\%), \textbf{EfficientNet-B0 (98.6\%)}, VGG-19 (96.86\%) \n
						\cite{electronics10121388} & 2021 & Plant Village (Whole) &  No  & ImageNet  & InceptionV3 (98.42\%), InceptionResNetV2 (99.11\%), MobileNetV2 (97.02\%) \& \textbf{EfficientNetB0 (99.56\%)} \n
						\cite{14844519620210101} & 2021 & Plant Village (Part) &   No & ImageNet & MobileNet (99.62\%) \& \textbf{InceptionV3 (99.74\%)}   \\
						\bottomrule
					\end{tabular} %
				}
				
				\footnotetext[1]{See Table \ref{Public_Dataset_Address}}

			\end{minipage}
			
		\end{center}
	\end{table}

Transfer Learning (TL) \JY{is a technique in machine learning that allows models trained on one task to be adapted to perform another task. It also} is a method to augment a learning model by reusing the knowledge learned from other domains for different (but related tasks). 
\JY{This could be useful in leaf disease classification, as models trained on one type of plant could potentially be adapted to work on other plants.} There are many related works in this direction, including domain adaptation and multi-task learning, however, in most practice, we can employ pre-trained models which are firstly trained from a huge, public dataset (e.g., ImageNet dataset) for other tasks, then deploy them on the target leaf disease dataset (e.g., Plant Village). In \cite{9355991},  the authors showed that through transfer learning the training time of CNN models can be shortened significantly. This idea has been deployed and studied widely in leaf disease classification. Table \ref{Transfer_Learning} lists the recent work about transfer learning methods in leaf disease classification.

A study in \cite{9231174} adopted several pre-trained deep learning models, including MobileNetV2, EfficientNetB0 and VGG-19, to classify tomato leaf diseases (1 healthy and 9 diseased classes). From the experimental results (MobileNetV2: 97.26\% accuracy, EfficientNet-B0: 98.6\% accuracy, VGG-19: 96.86\% accuracy), they claimed that transfer learning has several advantages:  smaller size models, less computational costs, and suitable on the mobile devices. In \cite{9355991}, the authors utilised a pre-trained VGG-16 and fine-tune their collected grape and apple leaves dataset. The model achieved 97.87\% accuracy, showing that through transfer learning CNN models' performance and efficiency can be improved. Another work in \cite{9293866} pointed out that one leaf may contain multiple leaf diseases in real life, thus, the authors used montage images to create the leaves which contain multiple diseases by combining nine pictures into one. Three pre-trained networks AlexNet, GoogLeNet \& ResNet-18 are tested, which achieved 95.65\%, 92.29\% and 89.49\% accuracy respectively. In \cite{9331214}, a pre-trained MobileNetV2 is used to classify 21 classes of healthy and diseased leaves (7800 images, resized to 224 * 224 pixels). Each class has 200 training samples, 100 validation samples and 50 test samples. The transferred  MobileNet can predict diseases with 90.38\% accuracy. \cite{15100606920210401} transferred pre-trained VGG-16, ResNet50 and InceptionV3 to classify rice leaf diseases. The dataset contains 3 leaf diseases and 1 healthy categories (resized to 224 * 224 pixels). Each class of the training set has 1000 images and each class of the test set has 300 images. Finally, the fine-tuned VGG-16, ResNet50 and InceptionV3 (with different hyper-parameters) achieved 87.08\%, 93.41\% and 95.41\% accuracy, respectively. \cite{9393311} deployed pre-trained GoogLeNet and VGG-16 for tomato leaf disease classification with accuracy of 99.23\% (GoogLeNet) and 98.00\% (VGG-16). A similar study can be seen in \cite{9278067} where the authors transferred a pre-trained VGG-16 to classify tomato leaf diseases. They tested several types of VGG-16, including (i) a fresh VGG-16 (training from scratch); (ii) a classic transfer learning VGG-16 pre-trained on ImageNet; (iii) a pre-trained VGG-16 with incorporated dropout and L2 regularization; and (iv) a pre-trained VGG-16 with dropout and an attention module. In the results, they claimed that the (iv) version with dropout operation and an attention module can effectively improve the accuracy and reduce validation loss, better than other versions. The proposed model in \cite{sharma2021detection} is based on pre-trained ResNet50. Only its last layer was fine-tuned and a Global average pooling layer was added with two 512-neuron dense layers on top. The result of this model, 98\% F1-score, shows the advantage of transfer learning. \cite{9138030} presented a pre-trained ResNet-50 with a data augmentation method to detect and classify 6 categories of tomato leaf diseases (Plant Village). The dataset was increased by four times through data augmentation. They showed that their proposed ResNet-50 model’s accuracy achieved 97\% after fine-tuning the transferred model. In \cite{electronics10121388} the authors transferred common pre-trained models InceptionV3, InceptionResnetV2, MobileNetV2, and EfficientNetB0 with depthwise separable CNN method to classify diseases in entire images of Plant Village dataset. The input size was set as 224 * 224 pixels. And they split the dataset into three test set types which are 20\%, 30\% and 40\%. Compare with other models, EfficientNetB0 gained the best accuracy of 99.56\% on the test set. They observed that different split types have little impact on this study. Using a smaller subset (5 types of crops from Plant Village) \cite{14844519620210101} tested fine-tuning MobileNet and InceptionV3 models. In this work, the leaf images were all processed by the segmentation method, and the two models achieved  99.62\% accuracy and 99.74\% accuracy, respectively.

\vskip 1cm
\noindent\rule{4cm}{0.8pt}
\textbf{Take-home Messages}\noindent\rule{4cm}{0.8pt}
\begin{itemize}
    \item [1.] Both data and model augmentation can help improve the performance and robustness of machine learning approaches for leaf disease classification. More attention can be seen in transfer learning where pre-trained models can be reused and augment the learning on leaf images. 
    \item [2.] Although data augmentation can be useful some researchers are skeptical about its effect. This is because some data augmentation methods (e.g., random cropping, colour transformation) can change the semantics of original images, which may create misleading images and reduce the performance of classification models \cite{8803793}.
    
    \item [3.] More attention is being paid to transfer learning, as can be seen in table \ref{Transfer_Learning} are satisfactory. This is reasonable as there are abundant pre-trained models on image data available for public use.
    
    \item [4.] There can be promising ideas for combining data augmentation and model augmentation. However, this study has not been addressed properly. We would encourage more studies in this direction.
\end{itemize}
\noindent\rule{12cm}{0.8pt}

\section{Applications}
\label{sec:apps}
In this section, we will review different leaf disease classification applications, from prototyping/lab-based products to commercialised software. We categorise the applications into: Web-based apps, Mobile apps, and Devices \& Hardware.
\begin{table}[h!]
\begin{center}
	\begin{minipage}{\textwidth}
		\caption{Various Applications}\label{Various_Applications}%
		\resizebox{\textwidth}{!}{%
			\begin{tabular}{@{}p{0.25\textwidth}p{0.12\textwidth}cp{0.14\textwidth}p{0.21\textwidth}p{0.3\textwidth}@{}}				
					\toprule
					Name/Paper & Type & Year & Working Area & Features  & Other  \\
					\midrule
					Rice Plant Leaf Disease classification \& Severity Estimation \cite{9342653} & Web \& WhatsApp & 2020 & Rice Leaf Diseases & CNN + HCI Rice Leaf Diseases\footnotemark[1] & \begin{tabular}{@{}l@{}}Accuracy 85.7\% \\ See Figure \ref{fig:Picture_9342653}\end{tabular} \n
					Plant Disease Identifier& Website & N/A &  Tomato \& Potato Leaf Diseases & Plant Village (Tomato \& Potato) &\url{https://cropify.herokuapp.com/} \n
					Tomato Leaf Disease Detection \& Classification \cite{9397001}  & Mobile App & 2021 & Tomato Leaf diseases &  CNN + Plant Village (Tomato)&  Accuracy 97\%  \n					

					Agriculture Field Monitoring \& Plant Leaf Disease classification\cite{9137805} & Mobile App & 2020 & Plant Leaf Diseases & CNN + Plant Village (Part)& \begin{tabular}{@{}l@{}} Accuracy 87.43\% \\ See Figure \ref{fig:Picture_9137805}\end{tabular}\n

					A Mobile-Based System \cite{agriengineering3030032} & Mobile APP & 2021 &  Plant Leaf Diseases & Leaves Data (website) + Plant Village (Part) \& CNN & Accuracy 94\%, See Figure \ref{fig:agriengineering3030032_1} \& \ref{fig:agriengineering3030032_2},  \url{https://github.com/ahmed-pvamu/Agro-Disease-Detector}  \n

					
					Agrio & Mobile App & N/A &  Plant Leaf Diseases & AI-based Alert
					System, Andriod \& iOS Support & \url{https://agrio.app/}\n
					CropsAI & Mobile App &  2020 &  Plant Leaf Diseases & 5 species (Corn, Wheat, Tomato, Soy beans \& Rice),with remote sensing technology \& iOS Support & \url{https://download.cnet.com/CropsAI/3000-20418_4-78607761.html} \n	
					
					Plants Disease Identification & Mobile App & N/A & Plant Leaf Diseases & iOS Support, Price: \$2.99& \url{https://apps.apple.com/us/app/plants-disease-identification/id1487234676}\n	
					Plantix & Mobile App & N/A & Plant Leaf Diseases & Cover 30 major crops, provides a community \& Andriod Support & \url{https://plantix.net/en/}
					\n
					
					Leaf Doctor & Mobile App & 2014 & Plant Leaf Diseases & Provides disease severity estimation \& iOS Support
					& See Figures \ref{fig:Leaf_doctor_1}, \ref{fig:Leaf_doctor_2} \& \ref{fig:Leaf_doctor_3}, \url{https://apps.apple.com/us/app/leaf-doctor/id874509900}\n
					Robotic Vehicle (Automated classification of Leaf Diseases) \cite{9198326} & Device & 2020 & Basil/Tulsi leaf Diseases & K-Means \& SVM + Basil/Tulsi leaves (Self) & See Figure \ref{fig:Picture_9198326_1} \& \ref{fig:Picture_9198326_2} \n
					Smart Glass (Real-Time Leaf Disease classification) \cite{9182146} & Device & 2020 & Tomato Leaf Diseases & YOLOv3 \& CNN + Tomato Leaves (Self) & \begin{tabular}{@{}l@{}} Accuracy 82.38\% \\ See Figure \ref{fig:Picture_9182146}\end{tabular}\n
					A Framework with Raspberry Pi Camera \cite{edssjs.8410D5A320210101} & Device & 2021 & Tree Leaf Diseases & Fuzzy Based Function Network + Tree leaves (Self)& Avg Specificity 80.66\%  \n
					A IoT handheld device\cite{edssjs.4D1044AE20210101} & Device & 2021 & Plant Leaf Diseases & Leaves Dataset (Self) + Plant Village (Part) \& MobileNet + CNN & Accuracy 96.88\% \n
					
					A Plant Disease classification Drone \cite{2b1bee20d14245e49c23b3dfe45cded2}& Device & 2022 & Plant Leaf Diseases & Plant Village (Whole) + Pre-trained EfficientNetV2-B4, a Quadcopter DJI Phantom 3 Drone & Accuracy 99.99\% \n	
					Agremo Drones Project & Device & N/A & Spot Disease \& Weed Infestations& Drones can automatically work on large-scale farms & \url{https://www.agremo.com/casestudies/using-drones-to-spot-disease-and-weed-infestations-in-sugar-beet/}\\

					\bottomrule
				\end{tabular} %
			}					
			\footnotetext[1]{See Table \ref{Public_Dataset_Address}}
			
		\end{minipage}
		
	\end{center}
\end{table}

\begin{figure*}[p!]
	\centering
    	\begin{subfigure}{0.31\textwidth}
    		\centering
    		\includegraphics[width=0.9\textwidth]{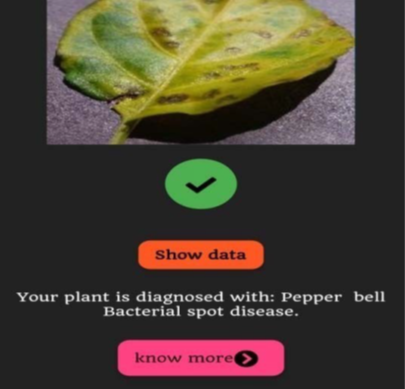}
    		\caption{ Prediction Result (Mobile App) \cite{9137805}}
    		\label{fig:Picture_9137805}
    	\end{subfigure}
    	\begin{subfigure}{0.3\textwidth}
    		\centering
    		\includegraphics[width=0.5\textwidth]{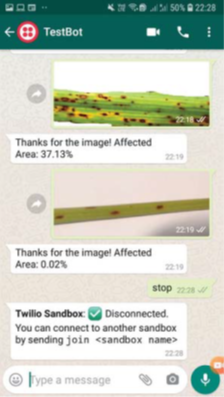}
    		\centering
    		\caption{Prediction Result (WhatsApp Interface) \cite{9342653}}
    		\label{fig:Picture_9342653}
    	\end{subfigure}
    \begin{subfigure}{0.31\textwidth}
    	\centering
    	\includegraphics[width=1.1\textwidth]{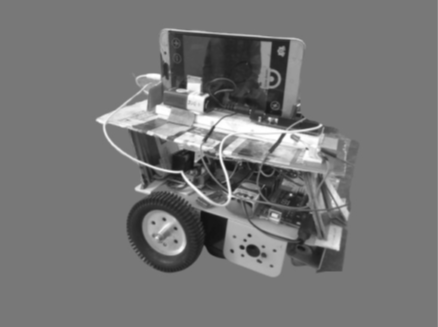 }
    	\centering
    	\caption{ The Robotic Car \cite{9198326}}
    	\label{fig:Picture_9198326_1}
    \end{subfigure}
    \begin{subfigure}{0.31\textwidth}
    	\centering
    	\includegraphics[width=1.1\textwidth]{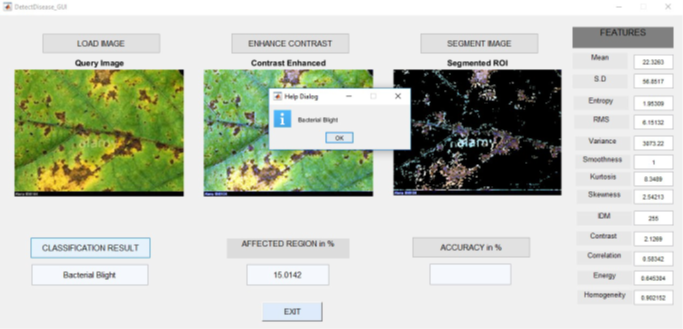}
    	\centering
    	\caption{Prediction Result (Robotic Car) \cite{9198326}}
    	\label{fig:Picture_9198326_2}
    \end{subfigure}
    \begin{subfigure}{0.31\textwidth}
    	\centering
    	\includegraphics[width=0.7\textwidth]{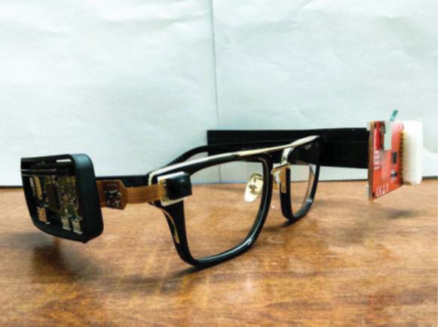}
    	\centering
    	\caption{Smart Glass \cite{9182146}}
    	\label{fig:Picture_9182146}
    \end{subfigure}
    \begin{subfigure}{0.31\textwidth}
    	\centering
    	\includegraphics[width=1.1\textwidth]{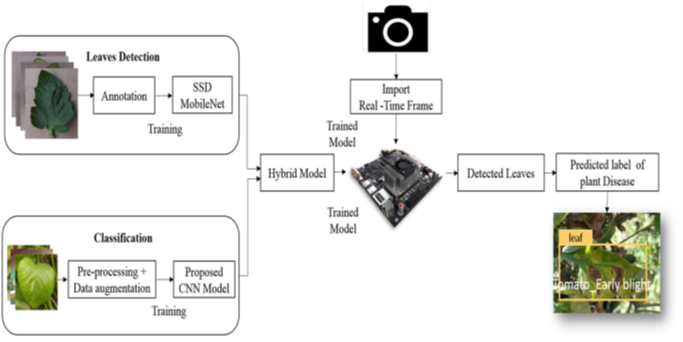}
    	\centering
    	\caption{The embedded platform \cite{edssjs.4D1044AE20210101}}
    	\label{fig:Picture_edssjs.4D1044AE20210101}
    \end{subfigure}

	\begin{subfigure}{0.45\textwidth}
		\centering
		\includegraphics[width=0.9\textwidth]{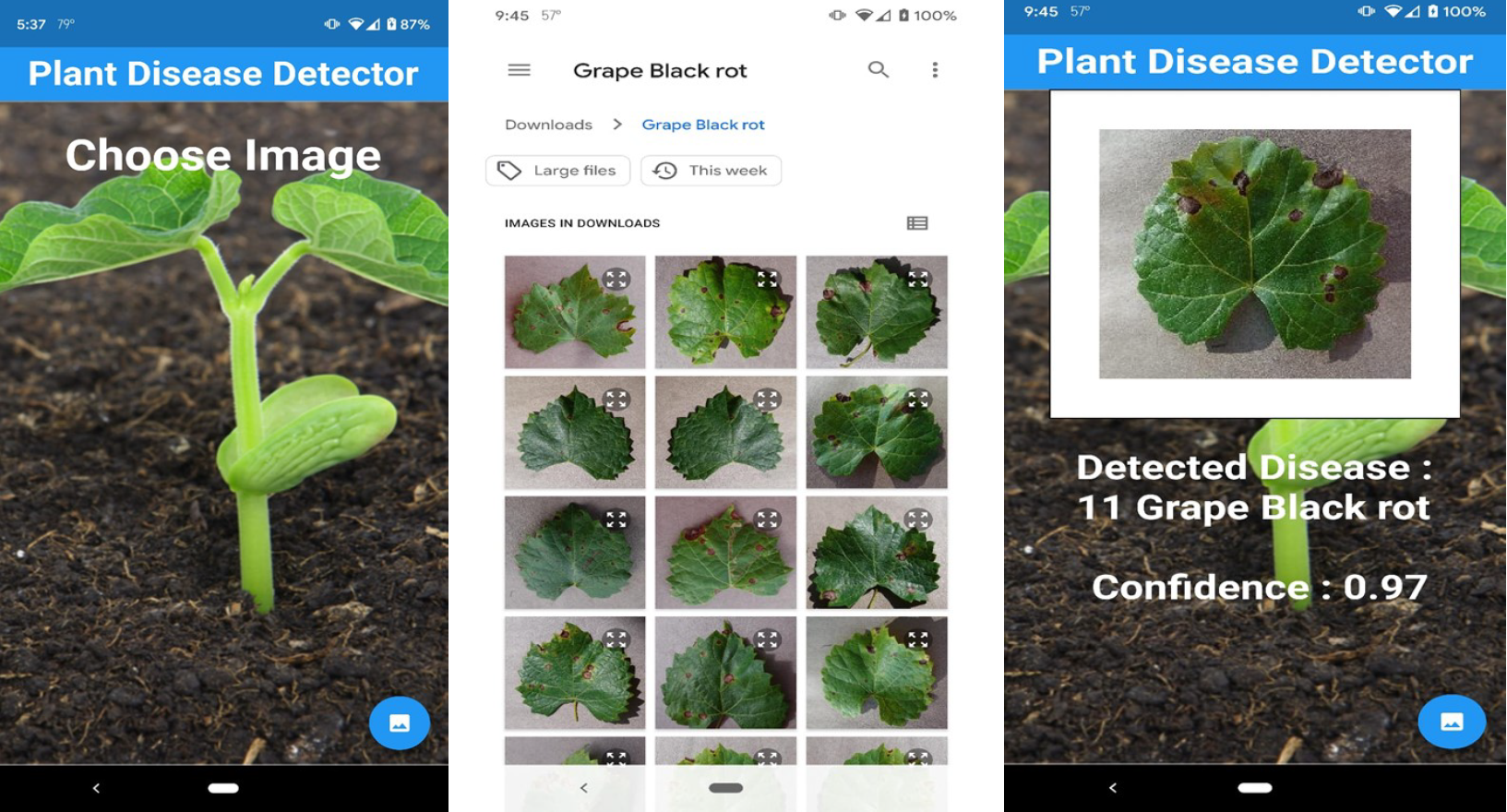}
		\centering
		\caption{ A Mobile-Based System (Interface) \cite{agriengineering3030032}}
		\label{fig:agriengineering3030032_1}
	\end{subfigure}					
	\begin{subfigure}{0.52\textwidth}
		\centering
		\includegraphics[width=0.9\textwidth]{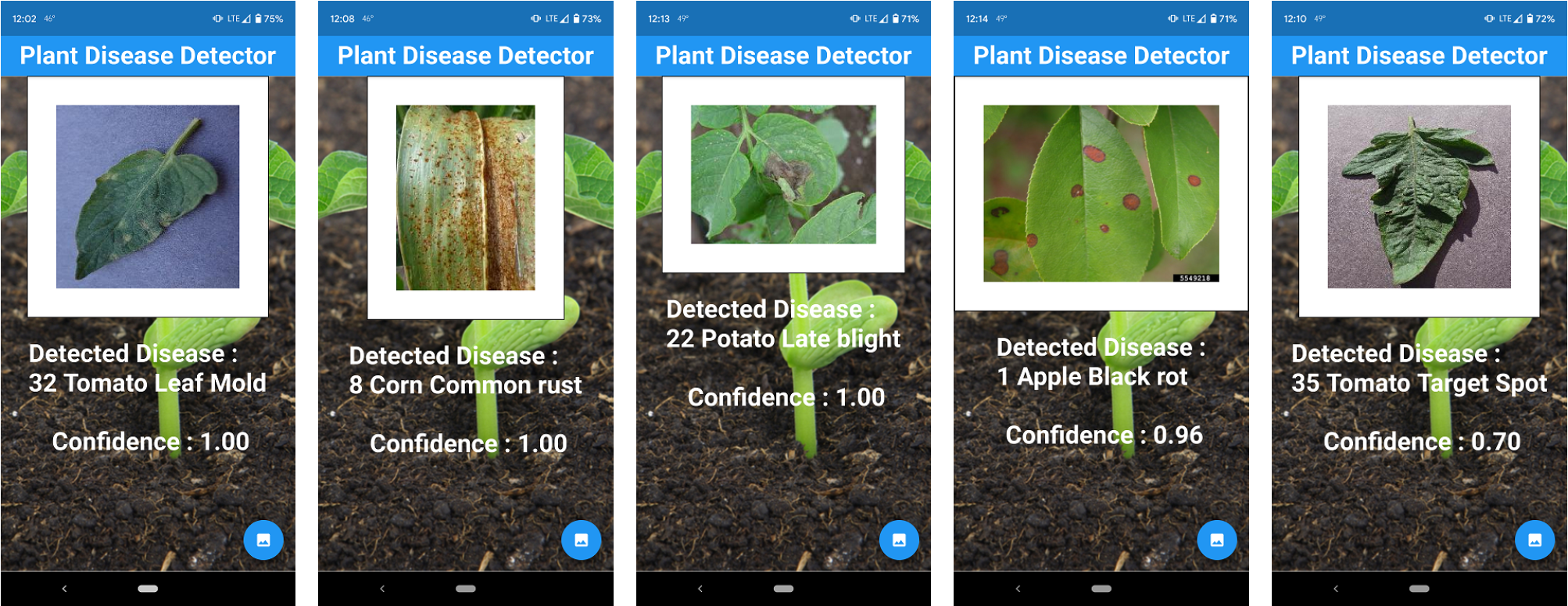}
		\centering
		\caption{ A Mobile-Based System (Results) \cite{agriengineering3030032}}
		\label{fig:agriengineering3030032_2}
	\end{subfigure}	
	
	\begin{subfigure}{0.31\textwidth}
		\centering
		\includegraphics[width=0.9\textwidth]{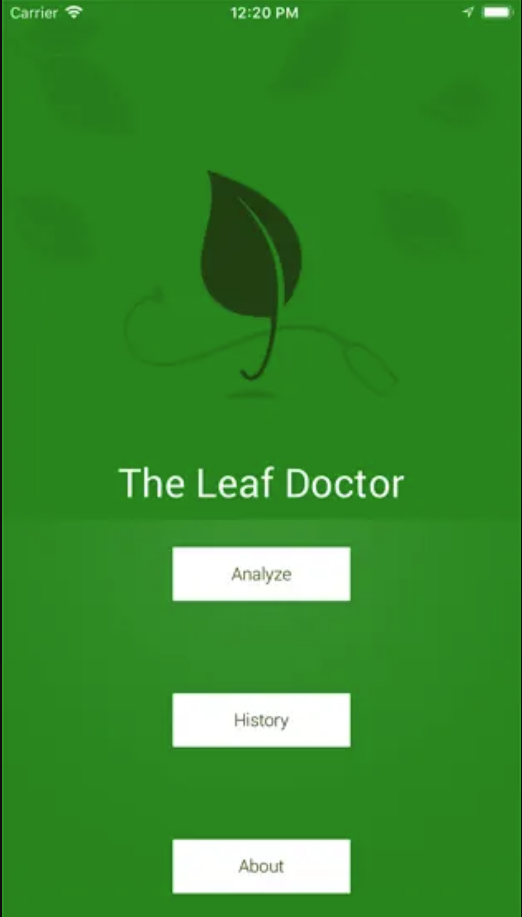}
		\centering
		\caption{Leaf Doctor Interface}
		\label{fig:Leaf_doctor_1}
	\end{subfigure}
	\begin{subfigure}{0.31\textwidth}
		\centering
		\includegraphics[width=0.9\textwidth]{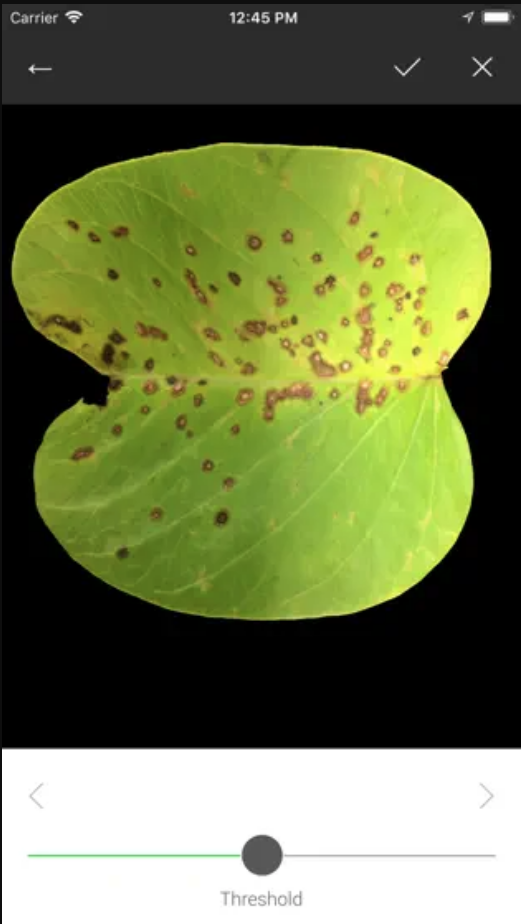}
		\centering
		\caption{Leaf Doctor (Selected) }
		\label{fig:Leaf_doctor_2}
	\end{subfigure}
	\begin{subfigure}{0.31\textwidth}
		\centering
		\includegraphics[width=0.9\textwidth]{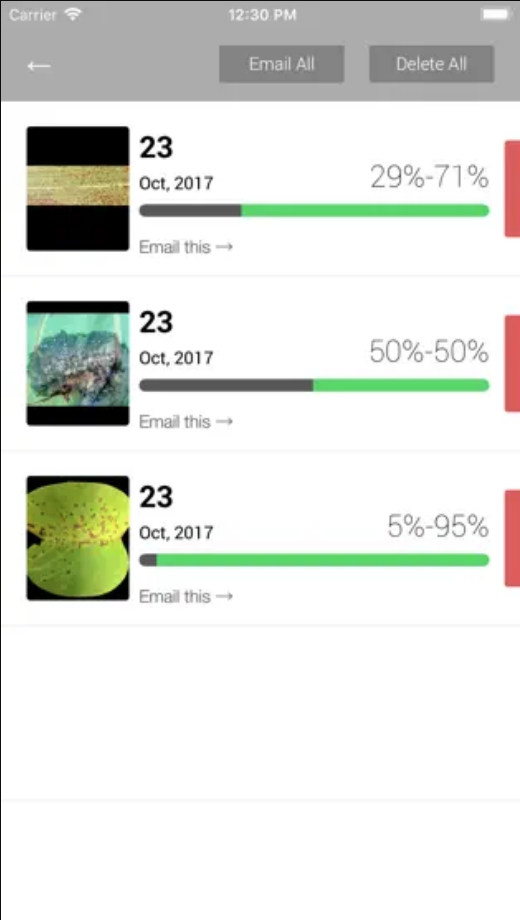}
		\centering
		\caption{Leaf Doctor (Results)}
		\label{fig:Leaf_doctor_3}
	\end{subfigure}						

	\caption{Various Applications of ML Technologies}
	\label{fig:Various_Applications_of_ML}
\end{figure*}

\subsection{Web-based Apps}
Website-based applications are always the first choice of industry or researchers because it is easy to use and not limited to hardware configuration. The user could submit a picture from a computer or a mobile phone, which was captured by a camera, to get predicted results in real time.

Several examples of web-based apps are shown in Table \ref{Various_Applications}. For example, Plant Disease Identifier (\url{https://cropify.herokuapp.com/}) is a website to provide tomato and potato leaf disease classification. A user only needs to choose a picture of the leaf to submit then will get the predicted result shortly. A rice disease classification system can be deployed on a website and WhatsApp (See Figure \ref{fig:Picture_9342653}). This system can diagnose three diseases of rice (based on a CNN model), and identify the severity of the diseased area (percentage, based on image segmentation). The dataset used here is the HCI Rice Leaf Diseases Dataset which contains 136 images of three rice diseases. The accuracy of this system is 85.7\% \cite{9342653}.

\subsection{Mobile Apps}
In recent years, mobile apps became more popular. Mobile apps can bring better user interface and user experience with the development and popularity of smartphones.

There are some examples of mobile apps for leaf disease classification from the industry. CropsAI is an iOS mobile app which can predict the common leaf diseases of 5 species (Corn, Wheat, Tomato, Soybeans \& Rice). Plants Disease Identification is a popular iOS mobile app with a price of \$2.99 on the App Store. Agrio is another mobile app which supports both Android and iOS. It claimed to have an AI-based alert system (needs remote sensors) that will notify the subscribed users and provide written preventative measures when detecting or expecting diseases or pests. Plantix is an Android mobile app which can classify leaf diseases of 30 main crops. It could provide instant disease classification and treatment advice. Notably, Plantix can have the largest online farmers and agricultural specialists community in the world \cite{agronomy12081869}. Users of Plantix could gain and share knowledge and help each other. Leaf Doctor was a mobile app created by the University of Hawaii, only available on the iOS system. Leaf Doctor supports leaf disease classification and provides disease severity estimation (See Figures \ref{fig:Leaf_doctor_1}, \ref{fig:Leaf_doctor_2} \& \ref{fig:Leaf_doctor_3}). The limitation of the mobile app is the software may be limited to smartphone systems and configuration. If a smartphone has a low configuration or outdated system, it will not work properly or will run the software slowly.

From the research community, both \cite{9137805} and \cite{9397001} designed a mobile app for leaf disease classification. The app in \cite{9397001} can classify tomato leaf diseases. Its training dataset was from tomato leaves of Plant Village and the prediction model was based on CNN. They showed that their app could achieve 97\% accuracy. Differently, the mobile app in \cite{9137805} can provide disease classification and real-time field factors monitoring (e.g., temperature, humidity, moisture) (See Figure \ref{fig:Picture_9137805}). It was based on a CNN model which was trained on part of the Plant Village dataset.  The authors demonstrated that their app can achieve 87.43\% accuracy on leaf disease prediction.

\subsection{Devices \& Hardware}
Devices or custom hardware are always required by professional agricultural specialists or researchers because the specific hardware can support more computing power and more reliable performance. In \cite{9182146} a study pointed out that existing deep learning approaches would need high processing power and may not be suitable for low-budget mobile devices. However, the high configuration will require more capital investment and professional technical capability requirements and training. We show some examples from research as follows. In \cite{9198326} a robotic vehicle was designed and developed (See Figure \ref{fig:Picture_9198326_1}) to detect Basil/Tulsi leaf diseases. Its components include a microcontroller, Bluetooth module, camera module and remote computer system. In the image detection module of the system, they used K-Means Clustering and SVM Classifier through MATLAB software. Users could get the prediction result from the software interface (See Figure \ref{fig:Picture_9198326_2}). In \cite{edssjs.8410D5A320210101} a novel framework (named \texttt{\detokenize{IoT_FBFN}}) was proposed. This framework is based on Fuzzy Based Function Network (FBFN) with IoT technology. It can capture real-time leaf images through the Raspberry Pi camera and transmit them to the system through the internet for FBFN network to classify diseases. They trained the system using a dataset of about 470 trees planted alongside the road in India. They demonstrated that rhe proposed system can achieve 80.66\% average specificity and 80.18\% average sensitivity, better than K-means and SVM. A handheld device (Embedded Platform) system was developed in \cite{edssjs.4D1044AE20210101}  (See Figure \ref{fig:Picture_edssjs.4D1044AE20210101}. With this handheld device, the classification accuracy rate can reach 96.88\%. The device will first detect leaves using a camera then divide the image and localise the leaves through data annotation and MobileNet. This module was trained on 338 leaf images they collected, 52 images online and 111 images from Plant Village. Finally, a custom CNN was used to classify diseases. This CNN was trained on 20 categories of Plant Village (apple, corn, potato \& Tomato). The system has a certain robustness capability against various conditions (e.g., weather, illumination \& background).  An interesting device, named Smart Glass, was developed in \cite{9182146}. This wearable device can be more convenient than the hand-held devices mentioned previously. It was based on a Raspberry Pi Zero W and can identify whether the leaf is healthy or not in real-time (See Figure \ref{fig:Picture_9182146}). The classification module used in Smart Glass is a transfer learning approach with YOLOv3 + CNN architecture fine-tuned on 304 tomato leaf images from farms (split into two categories: healthy and unhealthy). The proposed model can achieve an average accuracy of 82.38\%.

Besides hand-held and wearable devices, Unmanned Arial Vehicles (UAVs) are attracting more attention  \cite{agriengineering3030032, 2b1bee20d14245e49c23b3dfe45cded2}. UAVs have great potential in agriculture in the future. In \cite{2b1bee20d14245e49c23b3dfe45cded2}, a team designed a drone (quadcopter DJI Phantom 3) with pre-trained EfficientNetV2-B4 to detect leaf diseases. The classification module was trained on Plant Village and achieved near-perfect accuracy of  99.99\%. In the industry, American company Agremo started using drones to detect leaf diseases and weeds in sugar beet farms. Drones are especially suitable for continuous inspection and work on large-scale farms. They alleged their drones can provide plant counts, location data of certain weeds and diseases, or irrigation problem identification (water stress). The data of drones collected could produce data visualization easily for farmers analysing leaf diseases, weeds, water issues and so on.

\vskip 1cm
\noindent\rule{4cm}{0.8pt}
\textbf{Take-home Messages}\noindent\rule{4cm}{0.8pt}
\begin{itemize}
    \item [1.] A wide range of apps and devices have been built using machine learning techniques (mostly deep learning). 
    \item [2.] Mobile apps are becoming more popular than web apps for individual users thanks to their compactness and mobility. Meanwhile, UAVs (drones) have potential in large-scale farming. Some prototypes of hand-held and wearable devices were tested but they are not ready for commercialisation.
\end{itemize}
\noindent\rule{12cm}{0.8pt}
\section{Conclusions}
\label{sec:concl}
Despite machine learning techniques have been widely used in leaf disease classification, to our best knowledge, a comprehensive and up-to-date survey which can cover related available data, techniques and applications is still desired by the industry and research community. Therefore, in this paper, we surveyed about 100 recent related articles, collected and listed a series of public datasets which can be researched, analysed state-of-the-art machine learning approaches (i.e., shallow learning, deep learning \& augmented learning) and reviewed feasible applications in academia and industry. 
We have the following findings. In the data part, Maize Leaf (NLB) dataset could be the largest public dataset of single plant species at present while Plant Village is the most popular dataset. Plant Village, Plant Leaves and Plantae\_K are all laboratory datasets which can be useful for prototyping and evaluating machine learning models. However, real-field datasets, including PlantDoc would provide a more comprehensive evaluation and support for realistic applications. 
For technologies, shallow machine learning requires feature extraction from images \cite{9388488} to be useful for the disease classification task. The two most common methods are K-means clustering and grey-level co-occurrence matrix (GLCM), in which GLCM is more recommended. A combination of features is also encouraged, as it can help improve performance. Support vector machine (SVM) was the most common method for leaf disease classification in shallow machine learning. It is very suitable for both smaller (more likely to be linear) or non-linear datasets \cite{9261801}. Its better performance in comparison to other classifiers is evident in several studies. However, if suitable features are selected, KNN or RF also can achieve better accuracy.
Relative to shallow learning, Deep learning models have been proven useful and more effective than shallow learning for leaf disease classification which should be recommended in real-life applications due to their high accuracy. It is also more convenient as we can get rid of the feature extraction steps and minimise the manual effort for data processing. The common off-the-shelf deep learning models are CNN, AlexNet, VGG-16, ResNet, EfficientNet, Inception and MobileNet. Custom CNNs are highly encouraged as we should design an optimal model for different tasks. It was evident that custom CNNs perform better than off-the-shelf models. We can see that the datasets used in deep learning papers were relatively larger than in other studies. This is consistent with the fact that deep learning models are usually data-hungry. Most of the studies focus on the performance (accuracy) aspect of the task while a more comprehensive comparison with compactness and efficiency is still missing. There are a few papers that addressed these issues, for example, \cite{9057889} evaluates models' speed and  \cite{S187705092030690620200101} evaluates models' storage space.
Recent research proved that both data and model augmentation methods can help improve the performance and robustness of deep learning for leaf disease classification. More attention is on transfer learning where pre-trained models can be reused and augment the learning on leaf images. Although data augmentation can be useful some researchers are sceptical about its effect. The reason may be some data augmentation methods (e.g., random cropping, colour transformation) can change the semantics of original images, which may create misleading images and reduce the performance of classification models \cite{8803793}. The popularity of transfer learning is reasonable as there are abundant pre-trained models on image data (e.g., ImageNet) available for public use now.
For applications, section \ref{sec:apps} showed that a wide range of applications (software) and devices (hardware) have been built using machine learning techniques (mostly deep learning). Mobile applications are becoming more popular than web apps for individual users thanks to their compactness and mobility. Meanwhile, UAVs (drones) have advantages and potential in large-scale farming. Some prototypes of hand-held and wearable devices were tested but they may not be ready for commercialisation.
\JY{Last but certainly not least is the explainability of Machine Learning methods. With the increasing adoption of Machine Learning in the agriculture industry, there arises a pressing demand for models to be transparent and explainable. This may be important for enabling farmers to understand the decision-making process and trust this new technology method.}

Based on the above findings, we have the following suggestions. 
\begin{itemize}
	\item [1.] The available datasets listed are useful for domain-adaptation and multi-task learning, however, this is largely missing in the current literature.
	\item [2.] A machine learning model should learn from different datasets in a compositional manner where the model can effectively adapt to new tasks/datasets added in.
	\item [3.] For small datasets with a small set of disease classes, simple methods may achieve good results.
	\item [4.] Many studies use different experiment settings, including different partitions for training/validation/test which makes their results difficult to compare. Therefore, a benchmarking study is needed and encouraged.
        \item [5.] \JY{The research on explainability in this area remains worth attention, as the industry still requires a means to effectively explain decision-making by Machine Learning models to enable user understanding.}
	\item [6.] There can be a promising idea of combining data augmentation and model augmentation. However, this study has not been addressed properly. 
\end{itemize}

\section*{Declarations}
\textbf{Confict of interest} The authors declare that there is no conflict of interest.

\bibliography{main.bib}


\end{document}